\RequirePackage{fix-cm}
\documentclass[twocolumn]{svjour3}          
\smartqed  
\usepackage{graphicx}
\usepackage{amsmath}
\usepackage{siunitx}
\usepackage{tabu}
\usepackage{multirow}
\usepackage{array}
\usepackage{cite}
\usepackage{float}
\usepackage{gensymb}
\usepackage{hyperref}

\usepackage{subcaption}
\usepackage{graphicx}
\usepackage{booktabs}

\newcommand{\etal}{\textit{et al}.}

\usepackage{xcolor}
\usepackage{soul}
\usepackage{ulem}

\newcommand{\mzadd}[1]{#1}

\begin{document}

\title{Object-Based Visual \mzadd{Camera Pose Estimation} From Ellipsoidal Model and 3D-Aware Ellipse Prediction
}


\author{Matthieu Zins         \and
        Gilles Simon \and
        Marie-Odile Berger
}


\institute{Inria, Université de Lorraine, LORIA, CNRS\\
    \email{matthieu.zins@inria.fr, gilles.simon@loria.fr, marie-odile.berger@inria.fr}
}


\date{Received: date / Accepted: date}

\maketitle

\begin{abstract}
In this paper, we propose a method for initial camera pose estimation from just a single image which is robust to viewing conditions and does not require a detailed model of the scene. This method meets the growing need of easy deployment of robotics or augmented reality applications in any environments, especially those for which no accurate 3D model nor huge amount of ground truth data are available. It exploits the ability of deep learning techniques to reliably detect objects regardless of viewing conditions. Previous works have also shown that abstracting the geometry of a scene of objects by an ellipsoid cloud allows to compute the camera pose accurately enough for various application needs. Though promising, these approaches use the ellipses fitted to the detection bounding boxes as an approximation of the imaged objects. In this paper, we go one step further and propose a learning-based method which detects improved elliptic approximations of objects which are coherent with the 3D ellipsoids in terms of perspective projection. Experiments prove that the accuracy of the computed pose significantly increases thanks to our method. This is achieved with very little effort in terms of training data acquisition -- a few hundred calibrated images of which only three need manual object annotation. \mzadd{Code and models are released at \url{https://gitlab.inria.fr/tangram/3d-aware-ellipses-for-visual-localization}.}

\keywords{Visual Localization \and Pose from Objects \and Ellipse Prediction \and Ellipsoidal Model}
\end{abstract}



\begin{figure}[tb]
    \centering
    \includegraphics[width=.94\linewidth]{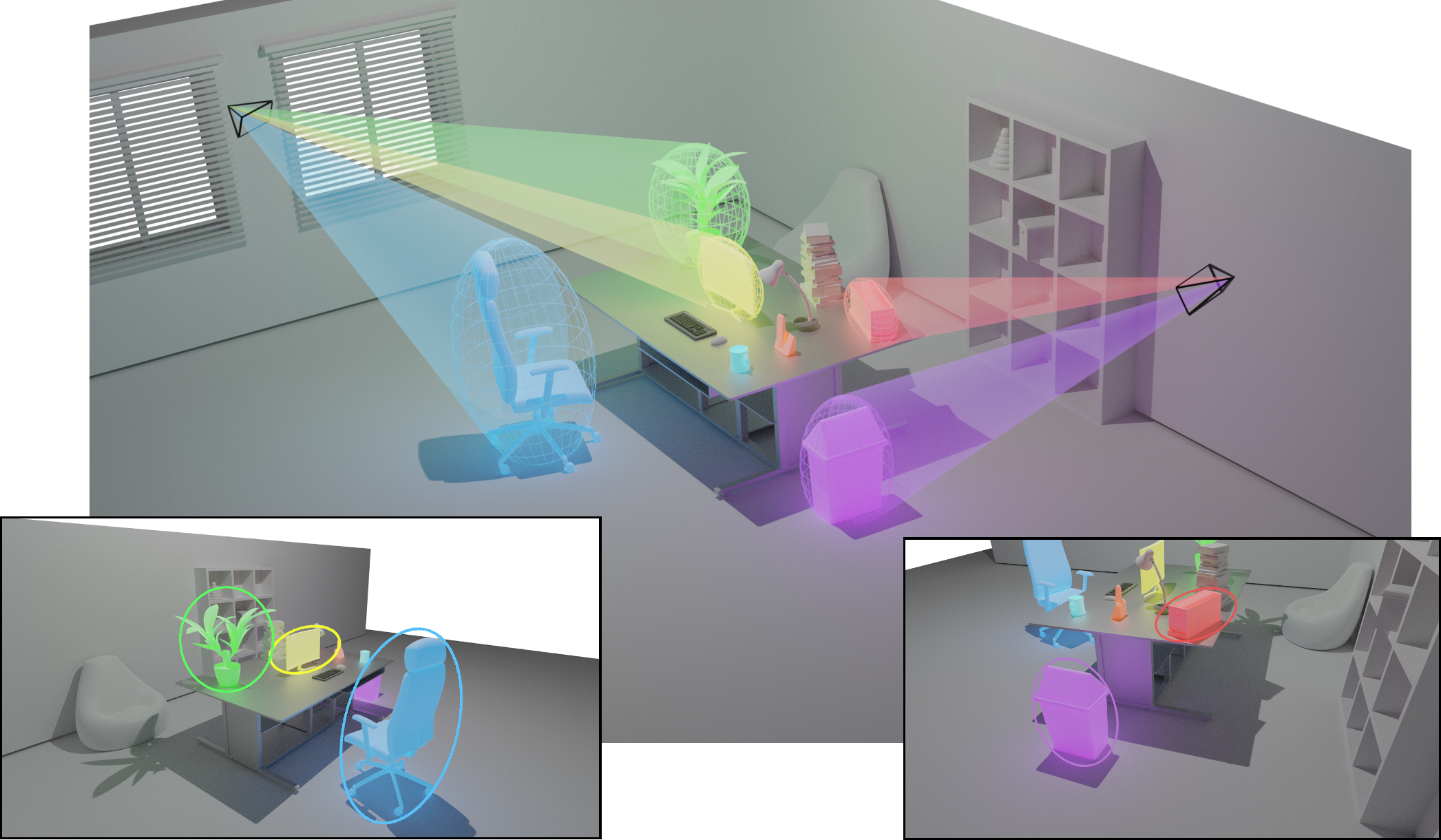}
    \caption{Camera pose estimation from objects.}
    \label{fig:Pose_from_objects}
\end{figure}


\begin{figure}[ht]
\centering
\begin{subfigure}{.48\linewidth}
  \centering
  \includegraphics[width=\linewidth]{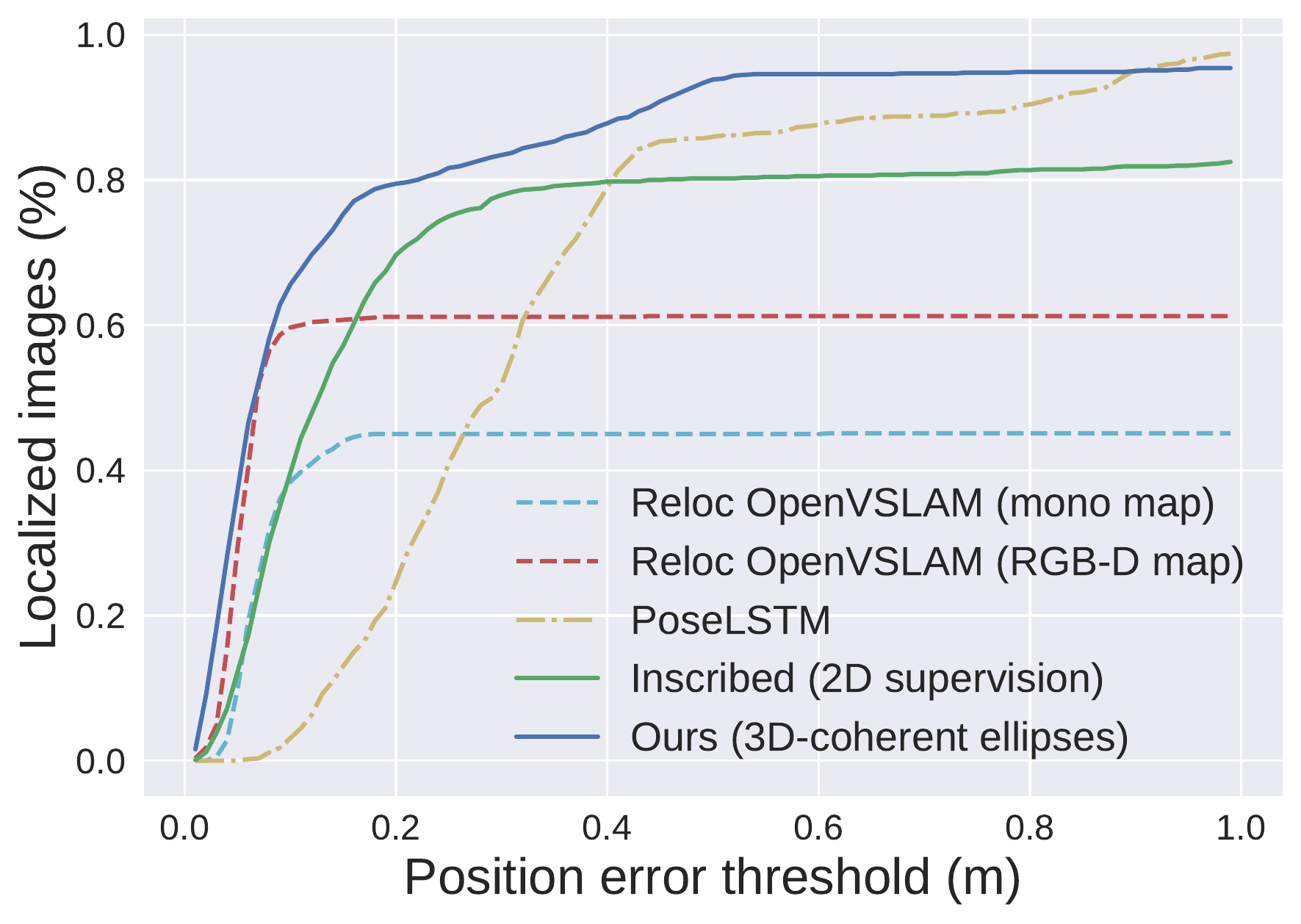}  
\end{subfigure}
\begin{subfigure}{.48\linewidth}
  \centering
  \includegraphics[width=\linewidth]{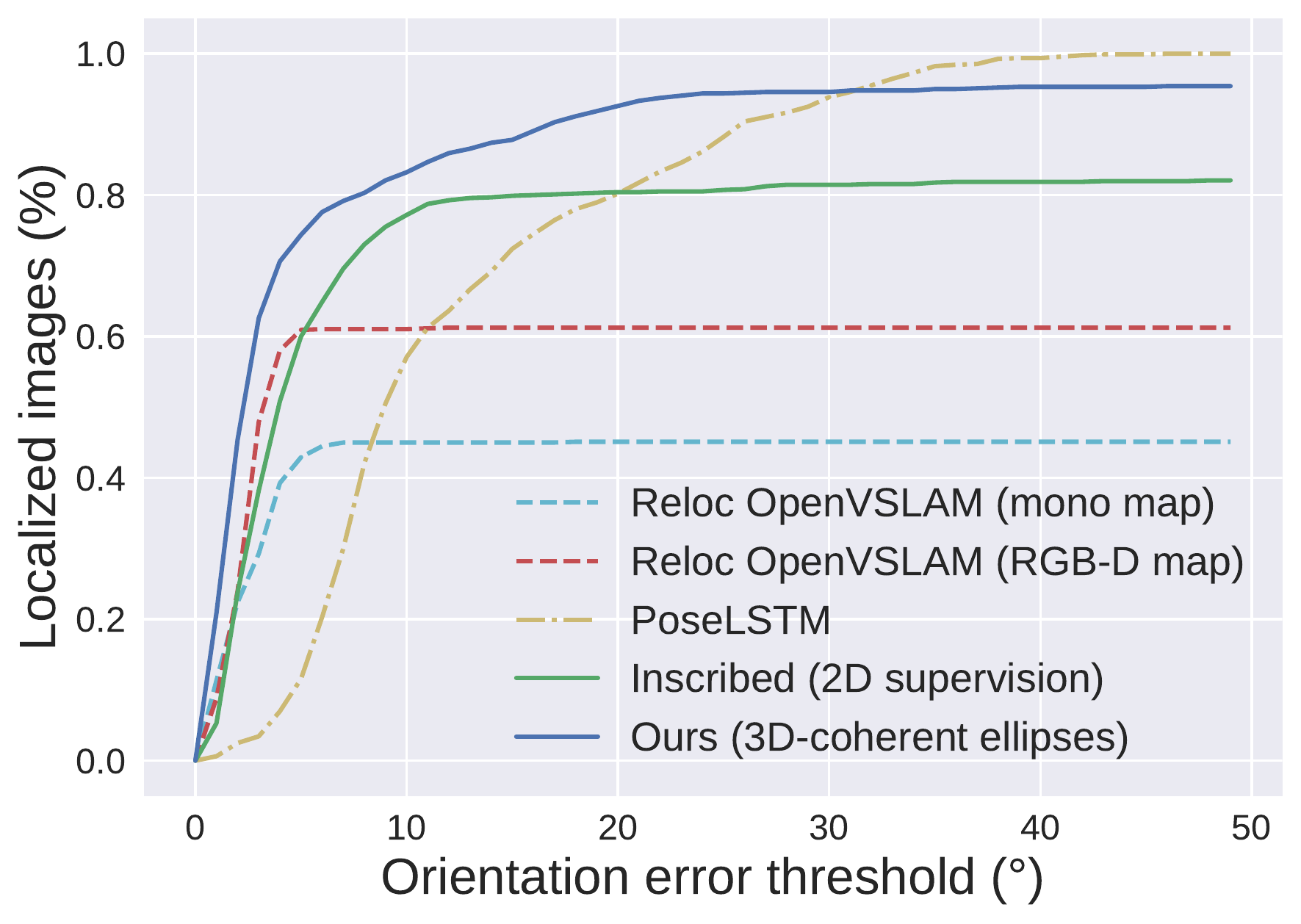}  
\end{subfigure}
\caption{Enhanced robustness of object-based visual localization compared to keypoint-based and direct pose regression methods.}
\label{fig:Enhanced_robustness_object_based_methods}
\end{figure}

\section{Introduction}
Estimating the 6-DoF pose of a camera from an RGB image is a fundamental task for Augmented Reality (AR) or robotics.
This task can be particularly challenging when no a priori knowledge of the pose is available, for example following a tracking failure or when trying to initialize the pose at the beginning of the process.


Classical methods generally rely on local descriptors and matching between the 2D keypoints detected in a query image and the 3D landmarks present in a previously reconstructed map point cloud~\cite{DBLP:conf/eccv/Sattler_active_search, DBLP:journals/pami/Sattler_loc, DBLP:conf/eccv/worldwide_pose}. The 6-DoF pose is then computed with the PnP algorithm inside a RANSAC loop. While these \mzadd{methods} can achieve very good accuracy, they usually require heavy computations and scene models. Also, they only provide a limited robustness to change of viewpoints or environmental settings, such as illumination.

New methods appeared with the advances of deep learning and, in particular, with the convolutional neural networks (CNN). They propose to directly regress the six parameters of the absolute camera pose using a CNN~\cite{KendallGC15,Kendall2016MUI,Melekhov2017ILU}. With these methods, the pose is obtained with a single forward pass in the network, rather than a heavy matching process. They also provide a better robustness to illumination changes. However, they do not reach the same level of accuracy as structured-based methods and have difficulties to generalize to viewpoints distant form the training images~\cite{SattlerZPL19}.

\mzadd{In this paper, we propose an object-based method for initial pose estimation from just a single image, which leverages the robustness of object detectors to large changes of viewpoints and environmental conditions. 
We model objects in 3D with ellipsoids, which can be interpreted as higher-level semantic landmarks for pose computation. A scene model made of ellipsoids is thus reconstructed in an initial step and is assumed to remain static for the localization. We use the term "object" here in a broad sense, i.e. any elements that can be detected by a specifically trained object detector. It is also conceivable to model larger objects by parts with multiple ellipsoids.
The major interest of the proposed method is its flexibility, provided by our rough ellipsoidal modeling that can be applied to any objects. In particular, we show that the fitting accuracy between the ellipsoidal model and the real object is not important.
Our method thus meets the growing need of easy deployment for robotics or augmented reality applications in any environments, especially those for which no accurate model nor huge amount of ground truth data are available.}




Many methods exist for estimating the pose of an object with respect to the camera frame~\cite{kehl2017, CDPN, PVNet,  RadL17, Sundermeyer2018, TekinSF18, Brachmann_6D_pose, DPOD, pix2pose}, however, they usually require a detailed textured model of the object and sufficient training images. This makes them unsuitable for our type of applications, as precisely digitizing all the objects and registering them in a global referential is challenging and not conceivable during the deployment on a new scene.


Similarly to~\cite{gaudilliere:hal-02170784,gaudilliere:hal-02886633}, we also use an ellipsoidal representation of the objects in the scene (Fig.~\ref{fig:Pose_from_objects}), which can be obtained with a coarse reconstruction.
Though these previous works are promising, the main source of inaccuracy originates from a poor approximation of objects in 2D with an ellipse aligned with the image axes and inscribed in the detection bounding box (BB).
Some methods exist to detect an object in the form of an ellipse. For example, Dong~\etal~\cite{dong2020ellipse} propose a direct elliptic detection and compare it with an ellipse fitting on the mask predicted by Mask R-CNN\cite{maskRCNN}, \mzadd{in the context of object 3D size and pose estimation}. In \cite{DBLP:journals/corr/ellipse_det_wacv}, the authors improve the detection of elliptic objects with the application of knots detection in sawn lumber images. However, \mzadd{this method is} dedicated to the 2D detection of elliptic shapes, but do not impose any projective coherency with a 3D model.
In this paper, we go one step further and propose a learning-based  method which detects improved  elliptic  approximations of objects which are coherent  with the 3D ellipsoid i.e. that are likely to be the projection of the ellipsoid. This way of detecting elliptic abstractions of objects significantly improves the accuracy of the recovered pose.
Our main contributions are as follows:
\begin{itemize}

\item A network for an improved 3D-aware object detection, which predicts ellipses around objects that are coherent with the projection of their 3D ellipsoidal abstractions. Its goal is to overcome the weaknesses of directly fitting the ellipses to the axis-aligned bounding boxes. Our data augmentation procedure allows for robustness to box boundaries variability. 



\item
We show how the concept of ellipsoidal abstractions of objects and 3D-coherent ellipse predictions can be used for robust pose computation when only a small amount of data is available on the scene. We show that the  pose accuracy  little depends on the choice of this ellipsoidal abstraction, which makes the method flexible and easy to use in practice. Only three calibrated images need to be annotated by hand to  build  the ellipsoid cloud. Annotations of the object are then obtained  by projection in  the training images.

\end{itemize}


This paper is an extended version of \cite{zins:hal-02975379}.  In this longer paper, we additionally provide:

\begin{itemize}
    \item A new loss formulation which handles more naturally the discontinuity of the angular parameter of an ellipse.
    
    \item Further investigations on the influence of the visible background in the crop images of the objects. \mzadd{We analyzed the benefits of using a ground truth object mask and proposed another masking strategy based on elliptic masks, which is more feasible in practice.}
    
    \item  A demonstration of the practical effectiveness of the method by exhibiting scenes where our object-based localization method outperforms the point-based relocalization module used in a state-of-the-art SLAM method (Fig.~\ref{fig:Enhanced_robustness_object_based_methods}).
\end{itemize}

\section{Background and related works}
\label{Related_Work}

Visual localization from monocular RGB images is an important problem in computer vision which witnessed a complete renaissance with the emergence of deep learning. Thanks to the ability of such methods to detect features across a wide range of viewpoints, largely independently from environmental conditions, this opens the way towards more robust localization and matching methods, especially able to handle few-textured scenes.

\subsection{Structure-based localization}
These traditional methods usually represent the scene as a point cloud and estimate the camera pose from 2D-3D matches between keypoints extracted from the query image and landmarks from the 3D map \cite{DBLP:conf/eccv/Sattler_active_search, DBLP:journals/pami/Sattler_loc, DBLP:conf/eccv/worldwide_pose}. This matching is generally based on local hand-crafted descriptors such as SIFT~\cite{DBLP:journals/ijcv/sift} or ORB~\cite{DBLP:conf/iccv/orb}, and the pose is computed using the PnP algorithm inside a RANSAC loop. However, these methods only succeed if enough points are correctly matched, which explains their relatively limited robustness to illumination changes, motion blur or large change of viewpoints.

More recent works leveraged the advances of deep learning to improve the keypoints detectors, descriptors and matching \cite{DBLP:conf/eccv/LIFT, DBLP:conf/cvpr/SarlinDMR20, DBLP:conf/cvpr/DeToneMR18}.

\subsection{Image-retrieval localization}
Image-retrieval methods estimate the camera pose from a query image by finding the most similar image in a database. They combine global descriptors (BoW~\cite{DBLP:conf/iccv/SivicZ03}, Fisher vector~\cite{DBLP:conf/cvpr/PerronninLSP10} or VLAD~\cite{DBLP:conf/cvpr/vlad1, DBLP:conf/mm/vlad2}), with efficient and scalable retrieval methods~\cite{DBLP:conf/cvpr/NisterS06, DBLP:conf/cvpr/PhilbinCISZ07}. With the emergence of convolutional neural networks, learned descriptors appeared~\cite{DBLP:conf/eccv/neural_codes}. The NetVLAD architecture was introduced in \cite{DBLP:conf/cvpr/NETVLAD} and showed remarkable results, outperforming the state-of-the-art non-learnt image representations and off-the-shelf CNN descriptors.

These methods can also be used as initial coarse pose estimation that is further refined. InLoc~\cite{DBLP:conf/cvpr/InLoc} combines image retrieval for large-scale initial pose estimation with dense matching for pose refinement. Piasco~\etal~proposed a fast and lightweight solution that combines image retrieval, dense matching and monocular depth prediction in~\cite{DBLP:conf/bmvc/PiascoSDG19}.

\subsection{Learning-based pose regression}
One of the pioneering method in the use of deep learning for pose computation is PoseNet~\cite{KendallGC15}, where the absolute camera pose is regressed using a CNN. By leveraging the notion of Bayesian networks, Kendall proposed a method for estimating the uncertainty of the predicted pose in~\cite{Kendall2016MUI}. A Long Short-Term Memory (LSTM) architecture was proposed by Walch~\etal~\cite{DBLP:conf/iccv/WalchHLSHC17} in order to address the problem of over-fitting. Kendall also replaced the original loss, which required hyper-parameters tuning, with a geometric learned loss in \cite{Kendall2017GLF}. These methods provided solutions to challenges for which classical methods failed, such as illumination changes or motion blur.
Also, they have a constant-time inference, compared to structure-based methods which often require heavy computations of 2D-3D matching inside a RANSAC loop. However, these methods have not yet reached the same level of accuracy and, as pointed out by Sattler~\cite{SattlerZPL19}, they are more closely related to pose approximation via image retrieval than to accurate pose estimation via 3D structure. As a result, such methods have difficulties to generalize to trajectories far from the training sequences.

\subsection{Scene coordinates regression}
These methods propose to regress dense 3D scene coordinates, originally with random forests~\cite{DBLP:conf/cvpr/Shotton_forest}, and more recently, by training a CNN~\cite{BrachmannR18, BuiAIN18}. The camera pose is then computed by solving a PnP problem, coupled with advanced versions of RANSAC~\cite{BrachmannDSAC}. These methods obtain remarkable results, but require depth information for training and are usually limited to small-scale scenes.

\subsection{Object-based methods}

Finding the pose of the camera from general shape objects can also be viewed as estimating the objects poses in the camera frame. Many works exist on this subjects \cite{kehl2017, CDPN, PVNet,  RadL17, Sundermeyer2018, TekinSF18, DPOD}. SSD-6D~\cite{kehl2017} extends the idea of 2D object detection and infers 6D pose based on a discrete viewpoint classification while an autoencoder is used in~\cite{Sundermeyer2018} to recover the object orientation. Another way to infer object pose is by predicting the 2D projections of the corners of the bounding box of the 3D object with a CNN. This avoids the need for a meta-parameter to balance the position and orientation error since the 6D pose can be estimated with PnP from 2D-3D correspondences. In BB8~\cite{RadL17}, segmentation is first performed to detect the objects and a CNN then infers the projection of the BBs. Data augmentation with a random background is performed during training to reduce the influence of the scene context. 

However, all these methods assume to have access to a detailed textured model of the objects. NOCS~\cite{NOCS} is an interesting category-level approach, in which a normalized coordinate space is used to represent different objects from a same category. Their large real and syntehtic dataset enable them to generalize to unseen objects from known categories. The objects poses are recovered by combining the predicted coordinate maps with a measured depth map.

While the above methods treat each object separately, in its own reference frame, other methods create maps of objects and localize the camera in it. Weinzaepfel~\etal~\cite{Weinzaepfel_2019_CVPR} propose a method where the camera pose is estimated from dense 2D-3D correspondences between the objects present in a query image and those in reference images. However, this method is limited to planar objects.
Yang~\etal~\cite{DBLP:journals/trob/cubeslam} integrated objects into a SLAM system by representing them with cuboids. In the context of autonomous driving,~\cite{MousavianAFK2017} also represent objects with 3D boxes and estimates their pose and dimensions using the geometric constraints provided by their 2D bounding boxes and some additional assumptions for their orientation. However, representing objects with 3D cuboids and the 2D detections with rectangles does not allow to derive closed-form solutions to the projection equations and leads to solutions with a high combinatorics. 

Modelling 2D/3D objects correspondences with ellipses/ellipsoids was already used by~\cite{rubino2018} in the context of multiview reconstruction and by Nicholson~\textit{et al}. in the context of SLAM~\cite{nicholson2019}. Resolution was based on the minimization of a geometric cost function between bounding boxes, using odometry sensors for initial position and orientation.

\mzadd{Recent works have proposed solutions for pose computation from ellipse-ellipsoid matching hypotheses without the need of an initial estimate~\cite{gaudilliere:hal-02170784}:} shows that the problem of estimating the camera pose from ellipse-ellipsoid correspondences has at most 3 degrees of freedom, since the position can be obtained from its orientation. Direct closed form solution can thus be estimated once the orientation is known. In~\cite{gaudilliere:hal-02886633}, a method for full pose recovery from at least 2 ellipse-ellipsoid correspondences was proposed under assumptions satisfied by many robotics applications. In practical experiments, axis-aligned ellipses are inferred from the bounding boxes detected by YOLO. The authors however note that such an elliptic 2D approximation is not always sufficiently accurate and may lead to a significant error on the estimated pose.

\section{Visual localization pipeline}

\subsection{Scene abstraction}
\begin{figure}[ht]
    \centering
    \includegraphics[width=\linewidth]{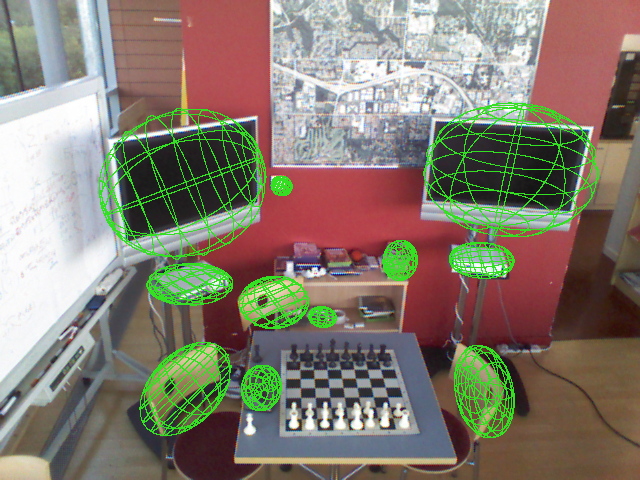}
    \caption{Reconstructed scene model for the \textit{Chess} scene.}
    \label{fig:7-Scenes_scene_model}
\end{figure}

In our method, we chose to represent our scene  with an ellipsoid cloud, where each object is simply modelled with one ellipsoid. Figure~\ref{fig:7-Scenes_scene_model} shows an example of such scene model, obtained on the \textit{Chess} scene.

\mzadd{This scene model is built once and does not evolve, as the goal of this work is to relocalize a single RGB image without using any temporal information.
The method used to build it is described in subsection~\ref{subsec:3D_model_and_training_data_generation}.}

\begin{figure}
    \centering
    \includegraphics[width=0.8\linewidth]{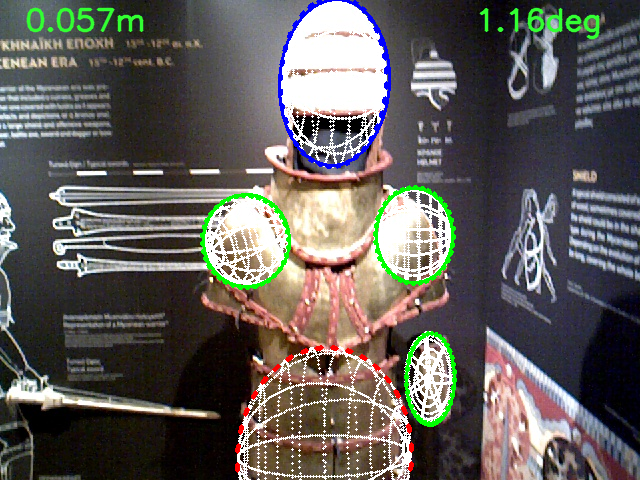}
    \caption{\mzadd{Camera pose estimated from a by-part modeling of a statue (head, left/right shoulders, left arm, bottom of the armor). The position error is written in the top-left corner and the orientation error in the top-right. The predicted ellipses are in solid line, the projections of the ellipsoids with the computed pose are in dashed lines and the white ellipsoids are projected with the ground truth camera pose. The green ellipses were used by P3P to compute the pose, the blue ellipse was considered as inlier in the validation step of the RANSAC and the red one was not used.}}
    \label{fig:result_MuseLearn}
\end{figure}

While being an approximate modelling of an object, this ellipsoidal representation has several advantages:
\begin{itemize}
    \item An object can be described with only 9 parameters (for the ellipsoid) with, potentially, one additional semantic attribute (i.e. the class of the object), which makes the scene model very compact and lightweight.
    \item The reconstruction of an ellipsoid from three ellipse observations has a closed-form solution, developed by Rubino~\etal~in~\cite{rubino2018}. For example, this would not be the case with 3D and 2D bounding boxes.
    \item \mzadd{With ellipsoidal objects and, contrary to what happens with 3D boxes, the equation of their projection~($C^*$) can be formally and continuously written as a function of the ellipsoid ($Q^*$) and the projection parameters ($P$): $C^* = PQ^*P^T$.}
    \item Ellipsoids were already used as primitives for decomposing objects, such as in~\cite{DBLP:conf/cvpr/PaschalidouUG19}. Even if treating an object by parts is not the focus of this work, the camera pose could also be estimated from such kind of decomposition. \mzadd{An example is given in Figure~\ref{fig:result_MuseLearn}.}
\end{itemize}

\subsection{Improved 3D-aware object detection}

In contrast to keypoint-based methods, where points have no shape but just a location, our objects have a shape both in 3D with the ellipsoid and in 2D with the ellipse.
As explained in~\cite{gaudilliere:hal-02170784, gaudilliere:hal-02886633}, computing the camera pose from pairs of ellipse-ellipsoid comes down to aligning their respective back-projection and projection cones. This requires a good coherency between our 3D abstractions of objects and their observations in the image. Ideally, the detected ellipse in the image should correspond to the intersection between the projection cone of the ellipsoid and the image plane.

Classical object detection methods usually predict axis-aligned bounding boxes~\cite{RedmonF17, FasterRCNN}. Some of them were also extended to predict a fine segmentation mask of the objects. However, all these methods are trained to perfectly fit to the objects contours, which is not coherent with our rough ellipsoidal models of objects.
On the one hand, an ellipse inscribed in an axis-aligned bounding box will fail to correctly represent an object as soon as it appears rotated in the image. On the other hand, a fine object segmentation mask would only be coherent with the projection of a perfectly detailed 3D model of the object, which can not be easily obtained in practice.

We thus propose to use an improved object detection method, with an ellipse prediction module specifically trained to be coherent with our scene modelling.

\subsection{Ellipse prediction}

\begin{figure*}
    \centering
    \includegraphics[width=\linewidth]{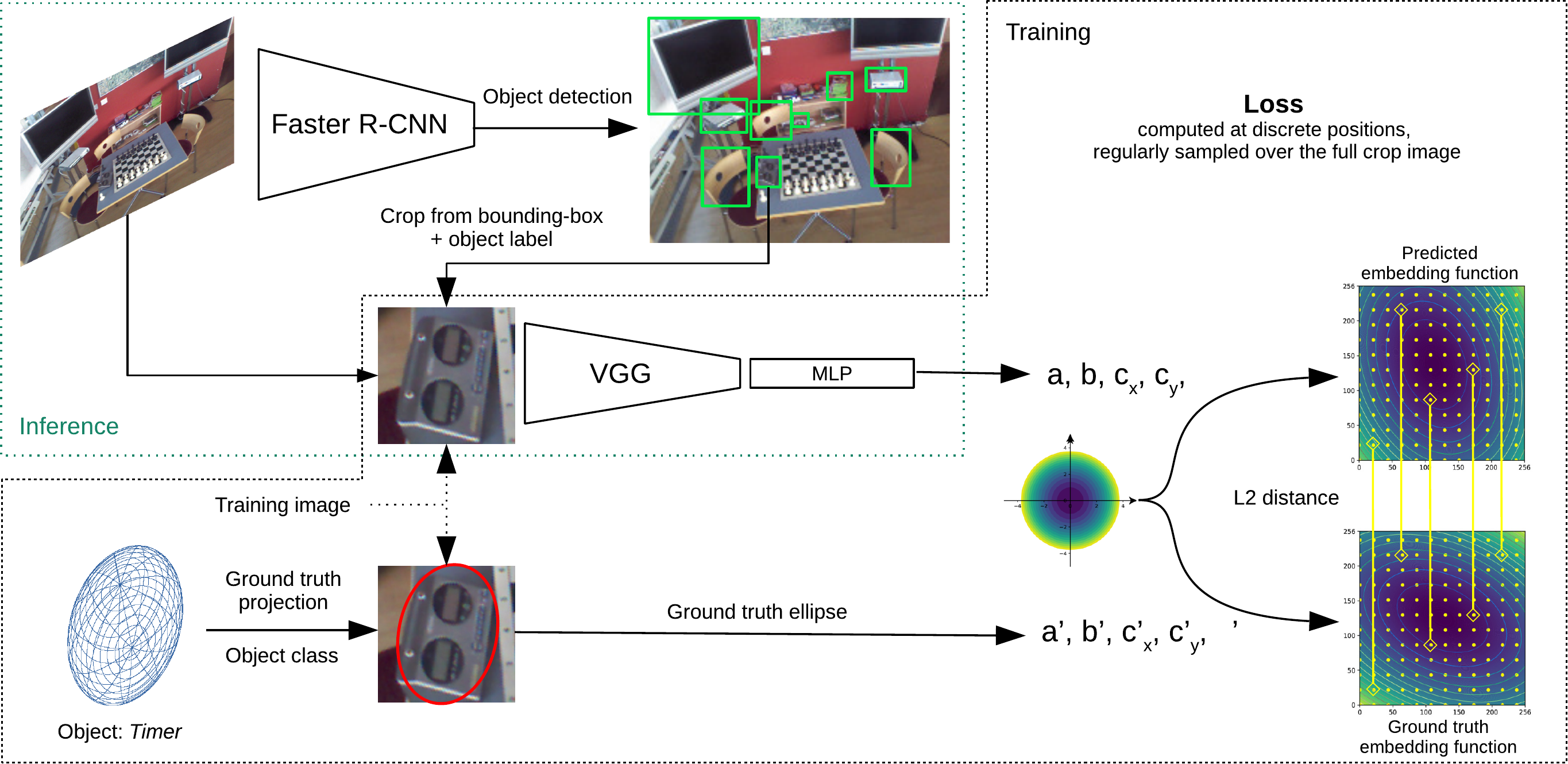}
    \caption{Proposed pipeline for training and inference. \mzadd{Here, the sampling grid was subsampled to $12\times12$ for the sake of visibility. In practice a grid of $25\times25$ points was used.}}
    \label{fig:System}
\end{figure*}

\paragraph{Ellipse parameterization.}
An ellipse is a special kind of conic which can be represented with the following quadratic equation:
\mzadd{
\begin{equation} 
\label{eq:ellipse_equation}
    (\mathbf{x}-c)^T R(\theta) \left[\begin{array}{cc}
        \frac{1}{\alpha^2} & 0 \\
        0 & \frac{1}{\beta^2}
    \end{array}\right] R(\theta)^T (\mathbf{x}-c) = 1
\end{equation}}
where $c$ is its center, $\theta$ its orientation and ($\alpha$, $\beta$) are the lengths of its semi-axes. The quadratic form of the ellipse can also be expressed as \mzadd{$\mathbf{x}^T C \mathbf{x}=0$} using homogeneous coordinates, in which the ellipse becomes a single symmetric $3\times3$ matrix. This matrix $C$ is defined up to a scale as the ellipse has only five degrees of freedom. However, although it could be possible to represent an ellipse with the five coefficients in the upper triangular part of this matrix, it is usually more convenient to use its physical attributes (position, size and orientation). Because of the symmetric nature of the ellipse, we always define the orientation as the angle between the horizontal axis and the part of the longest semi-axis which is in the right half of the ellipse. The possible values are constrained in the interval [$-\frac{\pi}{2}$, $\frac{\pi}{2}$].

\paragraph{Ellipse loss.}
Directly computing a loss between two ellipses using this representation is not straightforward. The axes and position parts can not be directly mixed with the orientation angle because of the discontinuity of the latter. Indeed, two very similar ellipses, just slightly rotated, can have a totally different orientation value (one at $\SI{89}{\degree}$ and the other at $\SI{-89}{\degree}$). To reduce this effect, a multi-bin approach with both a classification and a regression of the angular parameter was proposed in~\cite{zins:hal-02975379}.

\mzadd{To solution this problem of angular discontinuity, we propose here a new loss formulation in which an ellipse is represented by a 2D embedding function $\Phi : \Omega \subset \mathcal{R}^2 \to \mathcal{R}$.
The distance between the ground truth and the predicted ellipse is then defined between their respective embedding functions $\Phi_{gt}$ and $\Phi_{pref}$.}
\mzadd{
\begin{equation} 
    d^2(\mathcal{E}_{pred}, \mathcal{E}_{gt}) = \int_\Omega (\Phi_{pred}(\mathbf{x})-\Phi_{gt}(\mathbf{x}))^2 \mathbf{dx}
\end{equation}
}

\mzadd{
In practice, we measure this distance at discrete positions, sampled regularly over the whole input image passed to the network. We used a square grid of sampling with dimensions $25\times 25$.}

\mzadd{
\begin{equation}
    d^2(\mathcal{E}_{pred}, \mathcal{E}_{gt}) = \sum_{i=1}^N (\Phi_{pred}(\mathbf{x_i})-\Phi_{gt}(\mathbf{x_i}))^2
\end{equation}
}

\mzadd{
In the context of shape matching, a classical embedding function is the signed distance to the closest contour point~\cite{level-sets}:
}

\mzadd{
\begin{equation}
    \Phi(\mathbf{x}) = \left\{\begin{array}{ll}
        \mathcal{D}(\mathbf{x}, C) &\text{if} \; \mathbf{x} \; \text{inside} \; C \\
        \mathcal{-D}(\mathbf{x}, C) &\text{if} \; \mathbf{x} \; \text{outside} \; C \\
        0 &\text{if} \; \mathbf{x} \in C
    \end{array}\right.
\end{equation}}

\mzadd{
Computing the closest distance to a contour is not straightforward and, in our case of aligning two ellipses, simpler and more efficient functions can be used.
One of the most natural one is indeed the quadratic equation of an ellipse (Equation~\ref{eq:ellipse_equation}), representing it  as the level-curve of value 1. This equation defines an oriented non-isotropic distance map from the center of the ellipse.}



\mzadd{However, we observed numerical instability while training the network with this expression.
Huge values of gradients and strong irregularities in the loss can be noted in Figure~\ref{fig:New_loss_inverse_analysis} and can be explained by the expressions on the diagonal of the central matrix, $[\frac{1}{\alpha^2}, \frac{1}{\beta^2}]$ and their respective derivatives $[\frac{-2}{\alpha^3}, \frac{-2}{\beta^3}]$ which can become huge when $\alpha$ and $\beta$ are small.}

\mzadd{
We evaluated different forms for this central matrix, discussed in subsection~\ref{subsec:analysis_embedding_functions}, and finally simplified it with the following expression, which provides the better results:
}
\mzadd{
\begin{equation}
\label{eq:our_implicit_function}
    \Phi(\mathbf{x}) = (\mathbf{x}-c)^T R(\theta) \left[\begin{array}{cc}
        \alpha & 0 \\
        0 & \beta
    \end{array}\right]
    R(\theta)^T  (\mathbf{x}-c)
\end{equation}}

\mzadd{
Compared to the previous multi-bin loss proposed in~\cite{zins:hal-02975379}, this new loss has several advantages:
\begin{itemize}
    \item It combines all the parameters of the ellipse in order to avoid the arbitrary weighting that is usually necessary to compare different quantities (for example, distances and angles).
    \item It naturally handles the discontinuity of the angular parameter of the ellipse.
    \item It naturally handles the case of almost circular ellipses (undefined angle parameter).
\end{itemize}
}

\subsection{Network architecture}

\begin{figure}
    \centering
    \includegraphics[width=0.95\linewidth]{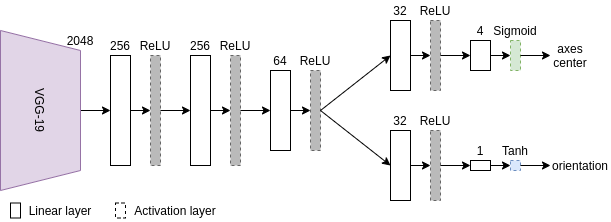}
    \caption{Network architecture for ellipse prediction.}
    \label{fig:Network_Architecture}
\end{figure}

The architecture of the neural network part of our system is described in Figures~\ref{fig:System} \mzadd{and \ref{fig:Network_Architecture}}. It mixes the standard Faster R-CNN architecture for object detection and a custom-designed network for the 3D-aware ellipse prediction.
\mzadd{
This second network takes as input a square subset of the image containing a detected object and resized to $256\times 256$ with a bicubic interpolation. The image crops are defined by the bounding boxes provided by Faster R-CNN and are forced to be square by using their largest dimension to avoid distortion.
}

This ellipse prediction network has a VGG-19 base followed by a few fully-connected layers, and finally, three branches predict the ellipse parameters: center (2 values), size (2 values) and orientation (1 value). The center and size are predicted with a final sigmoid activation layer so that their values are between 0 and 1. We interpret them as being normalized with respect to the size of the crop input image ($256\times256$).
\mzadd{The center and dimensions of the ground truth ellipses in the training data are thus also normalized.}


\mzadd{At test time, when the predicted ellipse is used for pose computation, the coordinates of its center are scaled back by $\frac{\max(w_{box}, h_{box})}{256}$ and translated by the coordinates of the top-left corner of the square detection box, in order to recover the coordinates in the original full-size image. The ellipse dimensions are scaled by the same factor.
}

As we defined the orientation angle in the right half of the ellipse, we can fully retrieve it from its sine value. The orientation branch thus ends with a hyperbolic tangent activation to produce a value between -1 and 1 that we interpret as the sine of the angle.

\subsection{Data association}

Similarly to keypoint-based methods, where a 2D-3D matching between image keypoints and landmarks is sought, we also need to associate the predicted ellipses with their corresponding ellipsoidal models \mzadd{available in the pre-built map. The process to reconstruct this map beforehand is explained in subsection~\ref{subsec:3D_model_and_training_data_generation}.} Also, the ellipse regression network is trained separately for each object of our scene model, and thus, the correct version of the network should be used for each detected object. We can only partly rely on the class label predicted by the object detector, because the scene might contain several instances of the same object class (Also, we can not leverage temporal consistency by using associated data from previous frames, as the goal of our method is to estimate the camera pose from a single image.)
Inspired by~\cite{gaudilliere:hal-02886633}, we use a robust RANSAC-based method, in which a score is computed for each association hypothesis. This score is computed using the object-wise Intersection-over-Union (IOU) between a detected ellipse in the image and the projection of its associated ellipsoid.

\subsection{Pose computation with ellipses-ellipsoids}

\begin{figure}[ht]
    \centering
    \includegraphics[width=\linewidth]{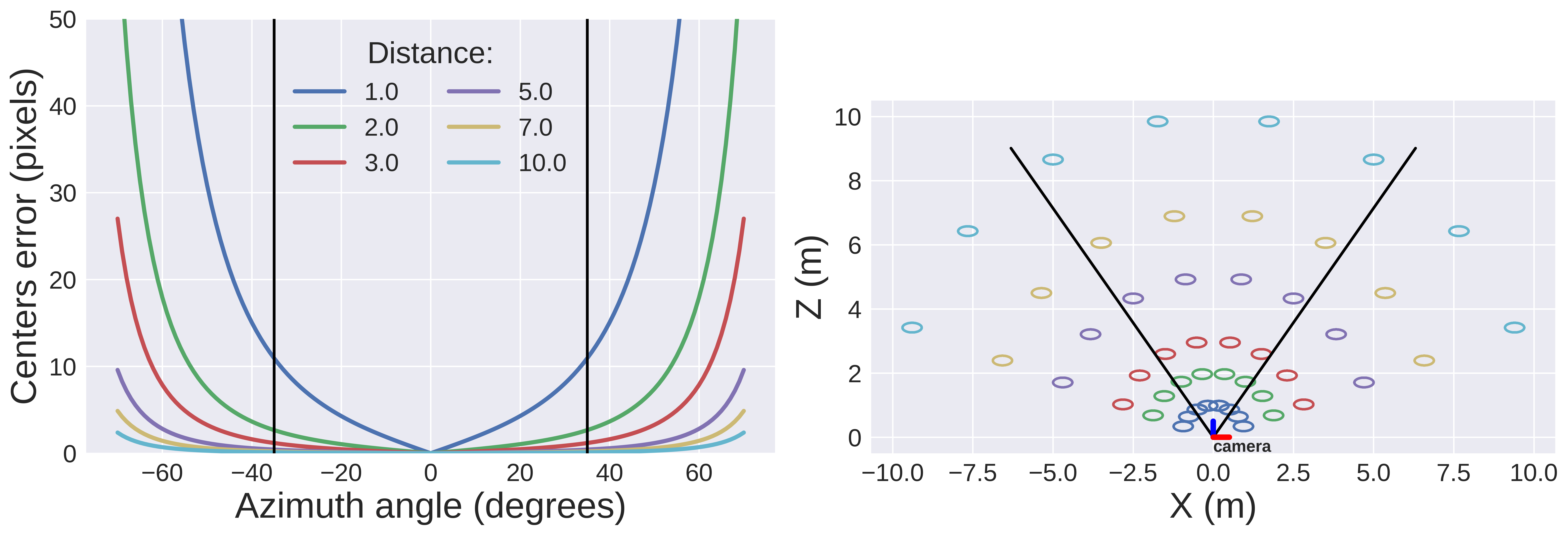}
    \caption{\mzadd{Distance between the center of the projected ellipse and the projection of the ellipsoid center. The experiment is illustrated on the right, where ellipsoids (of size [\SI{30}{cm}, \SI{20}{cm}, \SI{15}{cm}]) are placed at different azimuths and distances from the camera ($f_x=f_y=450$, image dimension: $640\times480$). In each figure, the black lines represent the field-of-view of a classical camera (70\degree).}}
    \label{fig:approximation_ellipsoid_center}
\end{figure}

RANSAC needs a direct method for pose computation from a minimal number of ellipse-ellipsoid correspondences. \cite{gaudilliere:hal-02886633} is the only work that describes a direct pose computation from two correspondences, but under the assumption of a near-to-zero camera roll. When the number of correspondences is larger, we used another strategy which consists in generating pose hypotheses from point-to-point correspondences between the ellipses and ellipsoids centers and validating them on the basis of a maximum IoU score. These strategies are described below:
\begin{itemize}
    \item When at least three objects are detected, the standard P3P algorithm between the ellipses and ellipsoids centers can be used. Assuming that the center of the ellipsoid projects on the center of the ellipse is wrong in theory, however, this is a totally realistic assumption in practice. The error remains quite small (only a few pixels, \mzadd{see Figure~\ref{fig:approximation_ellipsoid_center}}) in the field-of-view of a classical camera.
    
    \item When only two objects are detected, we use the P2E method described in~\cite{gaudilliere:hal-02886633}. Assuming that the camera roll is null, this method transforms the 6-DoF problem into a reduced problem with only one remaining degree-of-freedom which corresponds to an angular parameter. This makes it possible to review all the possible solutions in the same RANSAC that is used for data association.

    \item When only one object is detected, it is still possible to estimate the camera position if we have access to orientation data~\cite{gaudilliere:hal-02170784}. In practice, this can be obtained using an external sensor (IMU) or with an automatic vanishing point detection algorithm.
\end{itemize}

\section{Data acquisition and augmentation}
\subsection{3D model and training data generation}
\label{subsec:3D_model_and_training_data_generation}

The scene reconstruction and the generation of training data for our network only requires a set of calibrated images and a small amount of manual annotations. This makes the method easy to deploy in a new environment, which is very interesting from a practical point-of-view. The procedure works as follows:
\begin{enumerate}
    \item Choose a minimum of three images of the scene showing the object(s) from various viewing angles.
    \item Define boxes around the objects visible in these images and associate a label to each box.
    \item Fit ellipses to these boxes. Just taking the inscribed ellipses is sufficient here. \mzadd{Indeed, we show in the experiment in subsection~\ref{subseq:influence_of_the_reconstructed_ellipsoid} that the fitting accuracy between the ellipsoidal models and the objects in 3D has almost no influence.}
    \item Build the ellipsoid cloud which will be used as scene model.
    \item Reproject the ellipsoid cloud in all the training images to get ellipse annotations.
\end{enumerate}

The obtained ellipsoids obviously depend  on the  images chosen for reconstruction. Fortunately, we show in Section~\ref{subseq:influence_of_the_reconstructed_ellipsoid} that their size and orientation  may vary significantly without degrading the method performance.

\subsection{Data augmentation}
\label{ssec:data_augmentation}
Data augmentation plays an important role in the training of the ellipse prediction network and its generalization  with a relatively limited number of annotated images. Several strategies were performed during training:
\begin{itemize}
    \item Color jittering randomly changes the brightness, contrast and saturation of an image in order to simulate illumination changes.
    \item Blurring filters the images with a randomly-sized Gaussian kernel in order to accommodate different resolutions caused by the object distance.
    \item Shifting randomly translates the images so that the object is not always perfectly centered, which should accommodate noisy object crops.
    \item In-plane rotations as well as perspective deformations (homographies) were added to generate new views of the object. They can, for example, simulate a camera which is not held upright, or not aiming at the object center.
\end{itemize}

\section{Experimental results}
\subsection{Full camera pose estimation}
\label{subseq:7-Scenes}
We used the 7-Scenes dataset to evaluate our method for camera pose estimation. This dataset is a collection of seven indoor scenes scanned with an RGB-D camera. For each scene, several scanned sequences are provided with color and depth frames as well as ground truth pose annotations. We used the scene called \textit{Chess}, as it illustrates a typical environment where object-based methods can be used. We split the six available sequences as follows: sequences 1, 4, 6 for training and 2, 3, 5 for testing.

\paragraph{\mzadd{Training details.}}
\mzadd{We trained the ellipse prediction network for 100 epochs per object, with an initial learning rate of $5\times10^{-5}$, reduced by half after 50 epochs. The batch size was set to 16 and the Adam optimizer~\cite{Adam} was used.
The object detection network, Faster R-CNN, was fine-tuned on the objects of the scene, separated in seven categories (tv, xbox, chair, ...), for 2000 iterations with a base learning rate of $2.5\times10^{-4}$.}

\paragraph{Comparison with other methods.}
We tested two other visual localization methods, one using a classical point-based approach (OpenVSLAM) and a second one which directly regresses the camera pose with a trained network (PoseLSTM).
For OpenVSLAM, we built the map using the complete SLAM system on the training sequences 1, 4 and 6. We actually built two maps, one with the RGB-D SLAM and the second one with the monocular version (with a manually estimated scaling factor). Their results are respectively named \textit{RGB-D map} and \textit{mono map} in Tables~\ref{tab:7-Scenes_results} and \ref{tab:7-Scenes_accuracies_big_table}.
For localization, we used only its \textit{relocalization} module. It combines image matching (BoW) and keypoints (ORB), but the tracking and the motion model are disabled. In practice, this module is used when the slam is lost and needs to relocalize itself from only the map and a single image, which is a typical example of where our system can be used. PoseLSTM was trained on the 3000 frames provided in sequences 1, 4 and 6 during 2000 epochs.
To evaluate the benefits of our 3D-aware object detection, we also reported the results obtained with only the object detector part (for predicting bounding boxes) trained with a 2D supervision provided by manual annotations. The inscribed ellipses are extracted from the detection boxes and the same RANSAC procedure is used to estimate the camera pose.

\paragraph{Results.}
Table~\ref{tab:7-Scenes_results} shows the results obtained on the three test sequences, but only on frames where at least two objects were detected. Otherwise, a direct comparison with the object-based methods is not totally fair. We nevertheless reported the results on all frames of the sequences in the left column of Table~\ref{tab:7-Scenes_accuracies_big_table}. 
Also, note that the two SLAM-based relocalization methods sometimes fail and do not provide any pose results. Their median position and orientation errors are thus only computed on the frames where they succeeded.
The proportions of valid estimations are also reported, in which a pose is considered valid when its position error is less than \SI{20}{cm} and its orientation error less than \SI{20}{\degree}.

More complete results are available in Table~\ref{tab:7-Scenes_accuracies_big_table}. They show, for each test sequence, how the proportion of correctly localized images evolves when increasing the error threshold. The columns correspond to results obtained on all the frames of each sequence, but also on subsets of frames (only those with at least 2 or 3 detected objects).

\begin{figure}
    \centering
    \includegraphics[width=\linewidth]{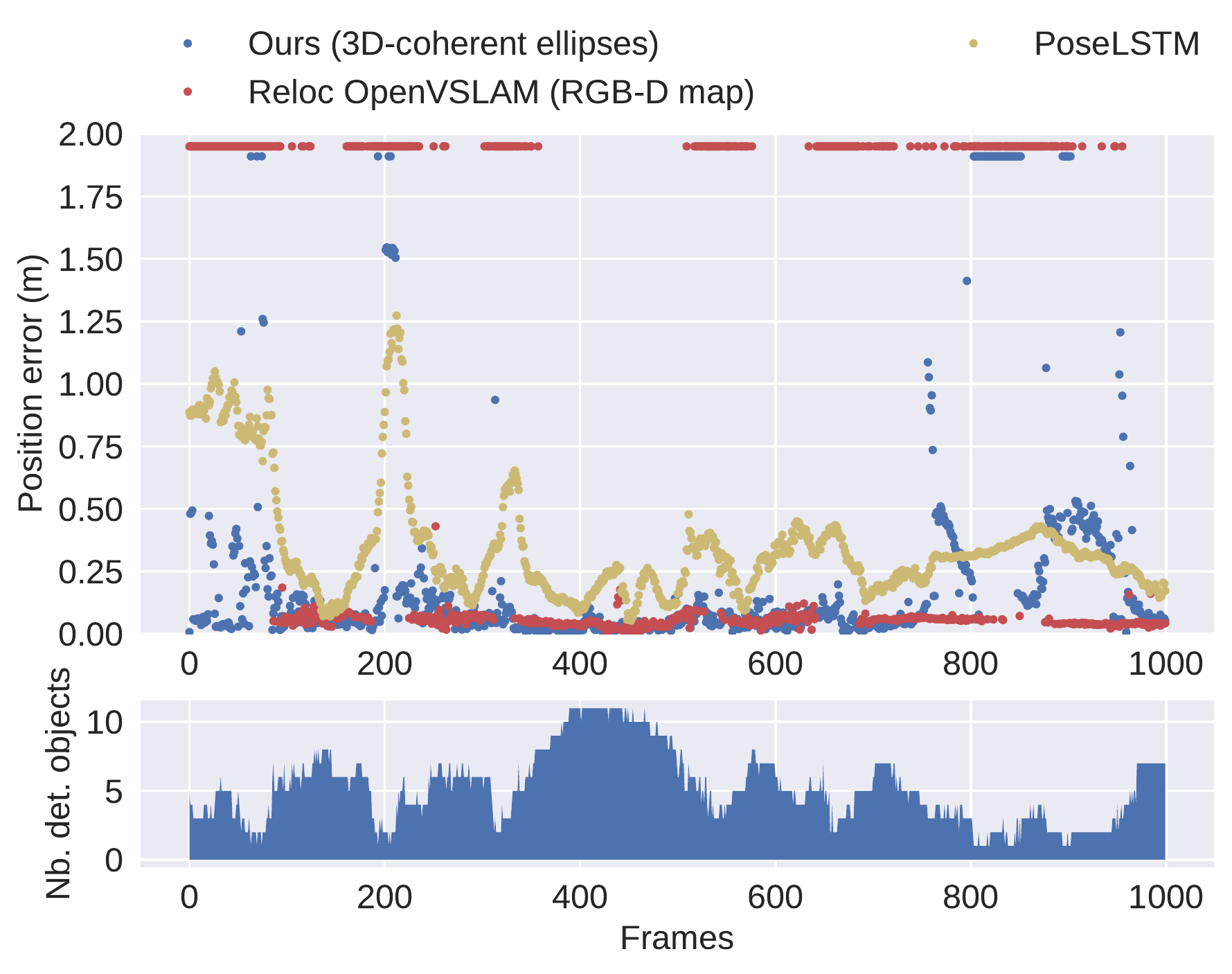}
    \caption{\textbf{Full camera pose estimation:} Position errors and the number of detected objects obtained on the frames of sequence 2 of the \textit{Chess} scene. The points above 1.75m correspond to frames where the method (either OpenVSLAM or our method) failed and could not provide any resulting pose.}
    \label{fig:7-Scenes_position_errors_analysis_seq_02}
\end{figure}

Figure~\ref{fig:7-Scenes_position_errors_analysis_seq_02} compares the position errors obtained on each frame of sequence 2. Note that the points above \SI{1.75}{m} correspond to frames where the pose estimation failed without returning any result (especially with OpenVSLAM and our method). For our method, this happens when strictly less than two objects are detected. In particular, the frames that failed (around 800-900) were taken with the camera very close to the table, and thus, only one (and sometimes two) object(s) could be detected.

The results clearly show the benefits offered by using objects as high-level landmarks. The keypoint-based method (with the RGB-D map) is slightly more accurate in position, but fails more frequently (especially in sequence 2).
PoseLSTM does not reach the same level of accuracy. However, it has the advantage being able to find a coarse pose for all the frames of the sequences. For example, in sequence 3, PoseLSTM can compute a coarse pose estimate for all the frames with an error not larger than \SI{65}{cm} in position and \SI{30}{\degree} in orientation.
Our method with 3D-coherent ellipses clearly outperforms the 2D-supervised and axes-aligned detections.

\begin{figure*}
    \centering
    \includegraphics[width=\linewidth]{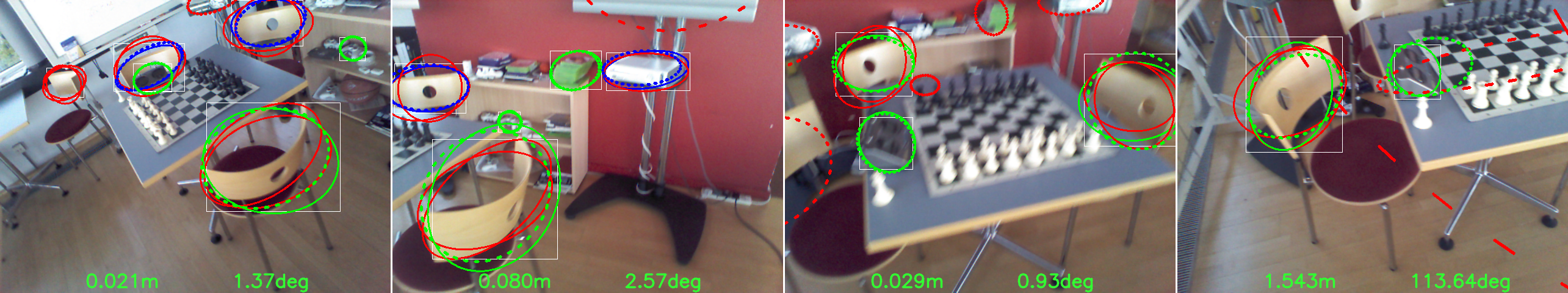}
    \caption{\textbf{Full camera pose estimation:} Some results obtained on frames from sequences 2, 3 and 5. In each image, the left value is the position error and the right value the orientation error. \mzadd{The detection boxes are in white, the predicted ellipses are in solid line and the projection of the objects ellipsoids with the estimated pose are in dashed lines.} The ellipses used for pose computation are in green, those considered as inliers in the validation process are in blue and the remainings are in red.}
    \label{fig:7-Scenes_some_results}
\end{figure*}

Figure~\ref{fig:7-Scenes_some_results} shows some localized frames. 
In particular, we can see the multiple ellipse hypotheses for the chair backs (as three instances exist in our scene model). The bold ellipses are the reprojections of the ellipsoidal models with the estimated pose. \textit{Green} stands for ellipses effectively used in the direct pose computation, \textit{blue} for ellipses considered as inliers in the validation process and \textit{red} for the unused ones. 
Note that, despite the fact that the left chair back detected in the first image was not in our scene model, the method is still able to find an accurate pose from the other objects.
The last image shows a failure case, which can happen when only two objects are detected in the image. Indeed, computing the pose with only two detections is particularly challenging as no other objects can be used in the IoU-based validation.
Finally, the third image shows the robustness of computing the camera pose from objects, despite the relatively strong motion blur.

\begin{table*}
  \scriptsize%
	\centering%
    \begin{tabu}{|c|ccc|ccc|ccc|}
  \hline
    \multirow{2}{*}{Method} & \multicolumn{3}{c|}{Sequence 2} & \multicolumn{3}{c|}{Sequence 3} & \multicolumn{3}{c|}{Sequence 5} \\
     & pos. err. & rot. err. & \% valid & pos. err. & rot. err. & \% valid & pos. err. & rot. err. & \% valid   \\

  \hline
   Reloc OpenVSLAM (RGB-D map)   &   \textbf{5.14}  & 2.41            & 61.15                & \textbf{5.05}& 2.53           & 77.93            & \textbf{4.57}   & 3.52          & 83.35 \\
   Reloc OpenVSLAM (mono map)    &   6.48           & \textbf{2.04}   & 45.0                 & 6.54         & 3.17           & 68.08            & 6.55            & 2.61          & 78.0 \\
   PoseLSTM                      &  29.15           & 8.94            & 24.69                & 18.73        & 6.06           & 53.68            & 16.16           & 6.03          & 62.28 \\
   Inscribed (2D supervision)    &  11.62           & 3.69            & 69.69                & 10.56        & 3.12           & 70.78            & 10.20           & 3.24          & 75.33 \\
   Ours (3D-coherent ellipses)   &   6.46           & 2.20            & \textbf{79.48}       & 7.03         & \textbf{2.12}  & \textbf{82.59}   & 6.42            &\textbf{2.05}  & \textbf{85.92} \\

  \hline
  \end{tabu}%
  \caption{\textbf{Full camera pose estimation:} Median position and orientation errors obtained on the \textit{Chess} scene (only on images with at least 2 objects detected). An estimated pose is considered valid when its position error is below \SI{20}{cm} and its orientation error below \SI{20}{\degree}.}
    \label{tab:7-Scenes_results}
\end{table*}

\subsection{Camera position estimation}
\label{subseq:Camera_position_estimation}
\begin{figure}
    \centering
    \includegraphics[width=\linewidth]{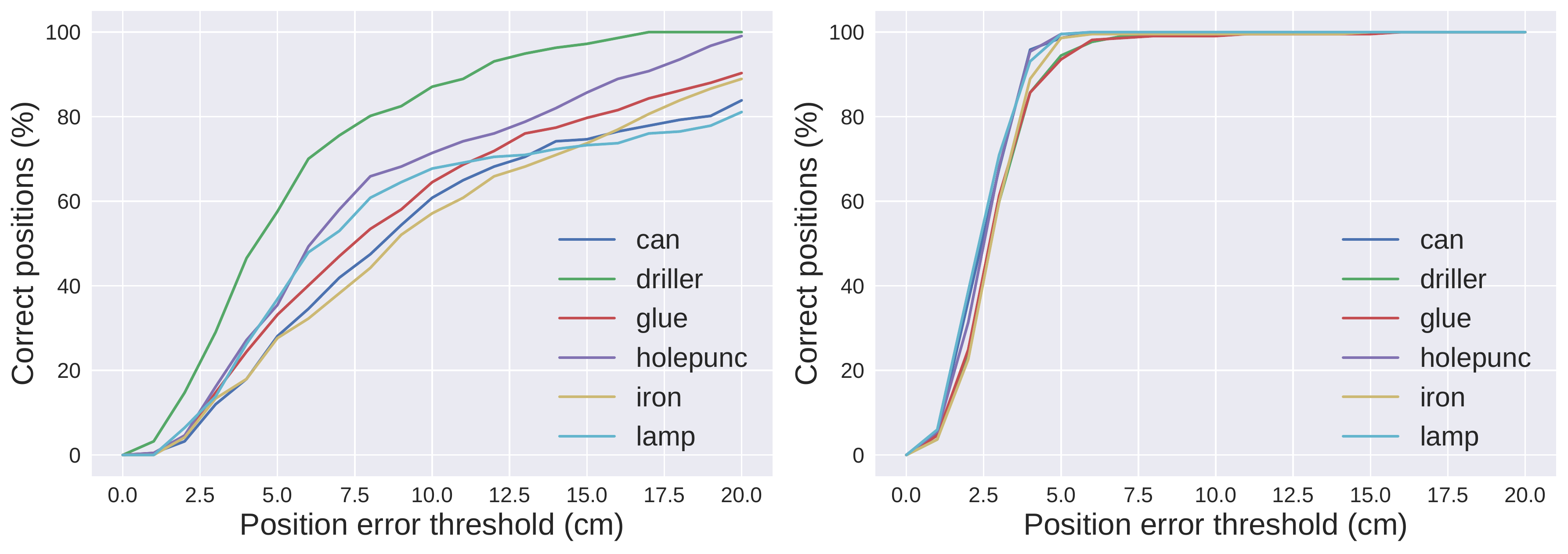}
    \caption{\textbf{Camera position estimation:} Proportion of camera positions correctly estimated  wrt. an increasing position error threshold. \textit{Left}: using the inscribed ellipse. \textit{Right}: using the predicted ellipse.}
    \label{fig:Linemod_position_error}
\end{figure}

\begin{figure}
    \centering
    \includegraphics[width=\linewidth]{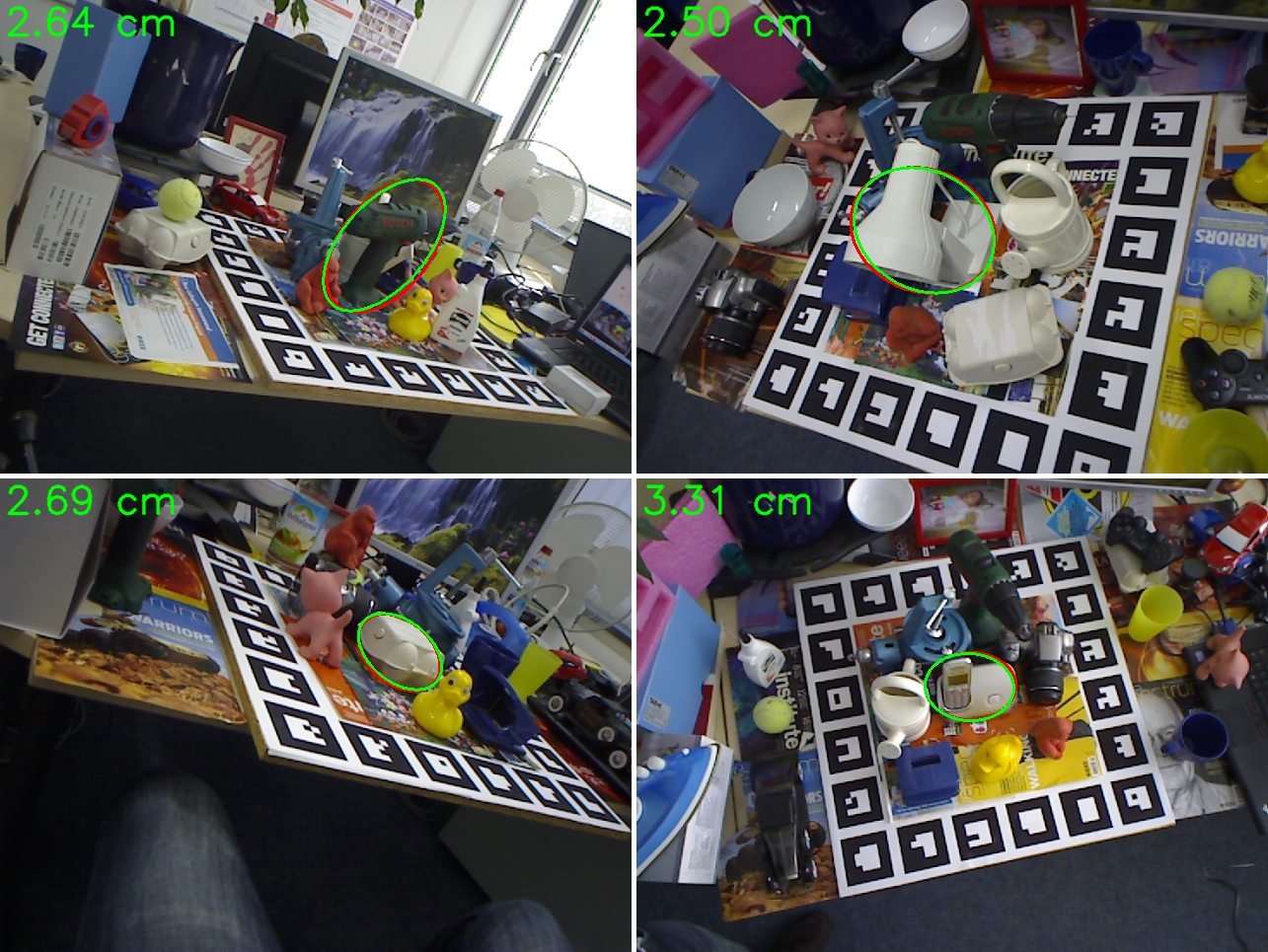}
    \caption{\textbf{Camera position estimation:} Predicted ellipse (in green) for some objects of LINEMOD. The red ellipse (partially behind the green one) corresponds the ground truth projection of the ellipsoidal object model. The reported value in each image is the error of the estimated camera position.}
    \label{fig:Linemod_predictions_for_some_objects}
\end{figure}

In some situations, it is possible to obtain the camera orientation using an external sensor (IMU) or a vanishing-point detection algorithm. In such scenarios, the camera position is determined from only one object. We evaluate here its accuracy and show the benefits offered by our improved object observation method compared to the axis-aligned ellipse. We used the LINEMOD dataset~\cite{HinterstoisserLIHBKN12}, which provides RGB-D images of 15 objects in cluttered environments with ground truth pose information. We split the available images in two, leading to around 200 images for testing and 200 for training. A few of them were used to reconstruct the ellipsoidal model of each object. Rather than assuming that we know the ground truth camera orientation at test time, we added a random noise, uniformly sampled in between $\SI{-2}{\degree}$ and $\SI{2}{\degree}$ on each of its Euler angle. This was done to be more realistic with what an external measurement could provide. Note that we did not fine-tune the object detection part of our system, but only evaluate the impact of the 3D-coherent ellipse prediction. Figure~\ref{fig:Linemod_position_error} and Tables~\ref{tab:Linemod_reproj}, \ref{tab:Linemod_ADD} show the proportion of correctly estimated positions wrt. the reprojection and ADD errors. These last two metrics compute the error between the object point cloud (provided in the dataset) transformed once with the estimated position and once with the ground truth value, either projected in the image (projection error) or directly in 3D~(ADD). These experiments show a significant improvement compared to the inscribed ellipses. Examples of predicted ellipses for some objects of the dataset are provided in Figure~\ref{fig:Linemod_predictions_for_some_objects}.

\mzadd{Finally, a comparison with previous methods for object 6D pose estimation is available in Table~\ref{tab:Linemod_ADD_other_methods}. Notice, that this comparison is given for information only, as our method does not use detailed 3D models of the objects, but assumes a known orientation when only one object is visible (a random noise uniformly sampled between -2° and 2° was added to each of the ground truth Euler angles).}

\begin{table}
  \scriptsize%
	\centering%
    \begin{tabu}{@{\hspace{0mm}}|@{\hspace{2mm}}c@{\hspace{2mm}}|@{\hspace{2mm}}c@{\hspace{2mm}}c@{\hspace{2mm}}c@{\hspace{2mm}}c@{\hspace{2mm}}|@{\hspace{2mm}}c@{\hspace{2mm}}c@{\hspace{2mm}}c@{\hspace{2mm}}|@{\hspace{0mm}}}
  \hline
 
    & \multicolumn{4}{@{\hspace{2mm}}c@{\hspace{2mm}}|@{\hspace{2mm}}}{} & \multicolumn{3}{c|}{}  \\
   Method & \multicolumn{4}{@{\hspace{2mm}}c@{\hspace{2mm}}|@{\hspace{2mm}}}{Inscribed ellipse~\cite{gaudilliere:hal-02170784}} & \multicolumn{3}{c|}{Ours}  \\
   & \multicolumn{4}{@{\hspace{2mm}}c@{\hspace{2mm}}|@{\hspace{2mm}}}{} & \multicolumn{3}{c|}{}  \\
   \hline
   Thresh. & 5 px & 10 px & 15 px & 20 px & 5 px & 10 px & 15 px\\
  \hline
      ape & 95.39 & 100.0 & 100.0 & 100.0 & \textbf{100.0} & 100.0 & 100.0  \\
      cam & 49.77 & 94.47 & 100.0 & 100.0 & \textbf{100.0} & 100.0 & 100.0  \\
      can & 57.60 & 79.26 & 98.62 & 100.0 & \textbf{100.0} & 100.0 & 100.0  \\
      cat & 68.20 & 98.62 & 100.0 & 100.0 & \textbf{100.0} & 100.0 & 100.0  \\
  driller & 16.13 & 61.75 & 90.32 & 98.62 & \textbf{96.31} & 99.08 & 100.0  \\
     duck & 89.40 & 100.0 & 100.0 & 100.0 & \textbf{100.0} & 100.0 & 100.0  \\
   eggbox & 97.70 & 100.0 & 100.0 & 100.0 & \textbf{100.0} & 100.0 & 100.0 \\
     glue & 54.38 & 88.02 & 95.85 & 99.54 & \textbf{100.0} & 100.0 & 100.0  \\
 holepunc & 83.41 & 100.0 & 100.0 & 100.0 & \textbf{100.0} & 100.0 & 100.0  \\
     iron & 17.05 & 51.15 & 78.34 & 93.09 & \textbf{98.16} & 99.54 & 100.0  \\
     lamp & 18.43 & 60.37 & 84.79 & 97.24 & \textbf{99.08} & 100.0 & 100.0  \\
    phone & 34.56 & 70.97 & 88.48 & 97.24 & \textbf{99.54} & 100.0 & 100.0  \\

  \hline
  \end{tabu}%
  \caption{\textbf{Camera position estimation:} Proportion of camera positions correctly estimated wrt. an increasing threshold of reprojection error, obtained on the LINEMOD objects.}
    \label{tab:Linemod_reproj}
\end{table}

\begin{table}
  \scriptsize%
	\centering%
  \begin{tabu}{@{\hspace{0mm}}|@{\hspace{2mm}}c@{\hspace{2mm}}|@{\hspace{2mm}}c@{\hspace{2mm}}c@{\hspace{2mm}}c@{\hspace{2mm}}|@{\hspace{2mm}}c@{\hspace{2mm}}c@{\hspace{2mm}}c@{\hspace{2mm}}|@{\hspace{0mm}}}
  \hline
 
     & \multicolumn{3}{@{\hspace{2mm}}c@{\hspace{2mm}}|@{\hspace{2mm}}}{} & \multicolumn{3}{c|}{}  \\
   Method & \multicolumn{3}{@{\hspace{2mm}}c@{\hspace{2mm}}|@{\hspace{2mm}}}{Inscribed ellipse~\cite{gaudilliere:hal-02170784}} & \multicolumn{3}{c|}{Ours}  \\
    & \multicolumn{3}{@{\hspace{2mm}}c@{\hspace{2mm}}|@{\hspace{2mm}}}{} & \multicolumn{3}{c|}{}  \\

   \hline
    Threshold & \multicolumn{3}{@{\hspace{2mm}}c@{\hspace{2mm}}|@{\hspace{2mm}}}{} & \multicolumn{3}{@{\hspace{2mm}}c@{\hspace{2mm}}|@{\hspace{0mm}}}{}\\

   (\% of diam.) & 10\% & 15\% & 25\% & 10\% & 15\% & 25\%\\
  \hline

      ape & 18.43 & 35.94 & 56.68     & \textbf{70.51} & 85.25 & 92.63 \\
      cam & 34.10 & 56.68 & 84.33     & \textbf{91.24} & 98.16 & 99.08 \\
      can & 12.90 & 18.43 & 31.34     & \textbf{93.09} & 97.70 & 98.16 \\
      cat & 26.27 &  37.33 & 55.30    & \textbf{76.04} & 92.63 & 96.77 \\
  driller & 42.86 & 57.14 & 76.04     & \textbf{85.71} & 92.17 & 97.24 \\
     duck & 31.34 & 47.00 & 67.28     & \textbf{66.82} & 83.87 & 92.17 \\
   eggbox & 16.59 & 22.58 & 40.09     & \textbf{88.48} & 94.47 & 97.24 \\
     glue & 11.98 &  23.04 & 32.72    & \textbf{70.51} & 82.95 & 92.63 \\
 holepunc & 12.90 & 20.74 & 30.88     & \textbf{86.64} & 92.63 & 97.24 \\
     iron & 16.59 & 25.81 & 40.55     & \textbf{91.71} & 98.62 & 99.54 \\
     lamp & 23.04 & 35.48 & 58.99     & \textbf{96.77} & 100.0 & 100.0 \\
    phone & 22.12 &  29.03 & 42.86    & \textbf{92.63} & 98.62 & 99.54 \\

  \hline
  \end{tabu}%
  \caption{\textbf{Camera position estimation:} Proportion of camera positions correctly estimated wrt. an increasing threshold of ADD error, obtained on the LINEMOD objects.}
    \label{tab:Linemod_ADD}
\end{table}

\begin{table}
  \scriptsize%
	\centering%
  \begin{tabu}{|@{\hspace{1mm}}c@{\hspace{1mm}}|@{\hspace{1mm}}c@{\hspace{1mm}}|@{\hspace{1mm}}c@{\hspace{1mm}}|@{\hspace{1mm}}c@{\hspace{1mm}}|@{\hspace{1mm}}c@{\hspace{1mm}}|@{\hspace{1mm}}c@{\hspace{1mm}}|@{\hspace{1mm}}c@{\hspace{1mm}}|}
  \hline
    \multirow{2}{*}{Method} & BB8 & U-D 6D & Tekin & Pix2Pose & Inscribed & \multirow{2}{*}{Ours} \\
      & \cite{RadL17} & \cite{Brachmann_6D_pose} & \cite{TekinSF18} & \cite{pix2pose} & ellipses \cite{gaudilliere:hal-02170784} &   \\
   
  \hline

      ape & 27.9 & 33.2 & 21.6 & 58.1 & 18.43  & \textbf{70.51} \\
      cam & 40.1 & 38.4 & 36.6 & 60.9 & 34.10  & \textbf{91.24} \\
      can & 48.1 & 62.9 & 68.8 & 84.4 & 12.90  & \textbf{93.09} \\
      cat & 45.2 & 42.7 & 41.8 & 65.0 & 26.27  & \textbf{76.04} \\
  driller & 58.6 & 61.9 & 63.5 & 76.3 & 42.86  & \textbf{85.71} \\
     duck & 32.8 & 30.2 & 27.2 & 43.58 & 31.34  & \textbf{66.82} \\
   eggbox & 40.0 & 49.9 & 69.6 & \textbf{96.8} & 16.59  & 88.48 \\
     glue & 27.0 & 31.2 & 80.0 & \textbf{79.4} & 11.98  & 70.51 \\
 holepunc & 42.4 & 52.8 & 42.6 & 74.8 & 12.90  & \textbf{86.64} \\
     iron & 67.0 & 80.0 & 75.0 & 83.4 & 16.59  & \textbf{91.71} \\
     lamp & 39.9 & 67.0 & 71.1 & 82.0 & 23.04  & \textbf{96.77} \\
    phone & 35.2 & 38.1 & 47.7 & 45.0 & 22.12  & \textbf{92.63} \\

  \hline
  \end{tabu}%
  \caption{\mzadd{\textbf{Camera position estimation:} Proportion of camera positions correctly estimated for an ADD error of 10\% on the LINEMOD objects.}}
    \label{tab:Linemod_ADD_other_methods}
\end{table}

\subsection{Robustness to new viewpoints}

One of the main limitation of existing methods for absolute pose regression is its low generalization ability to new viewpoints. 
Our previous experiment on 7-Scenes only partially evaluates this capacity, as the camera trajectories used to generate the training and testing images stay approximately in the same area.

\begin{figure}
    \centering
    \includegraphics[width=\linewidth]{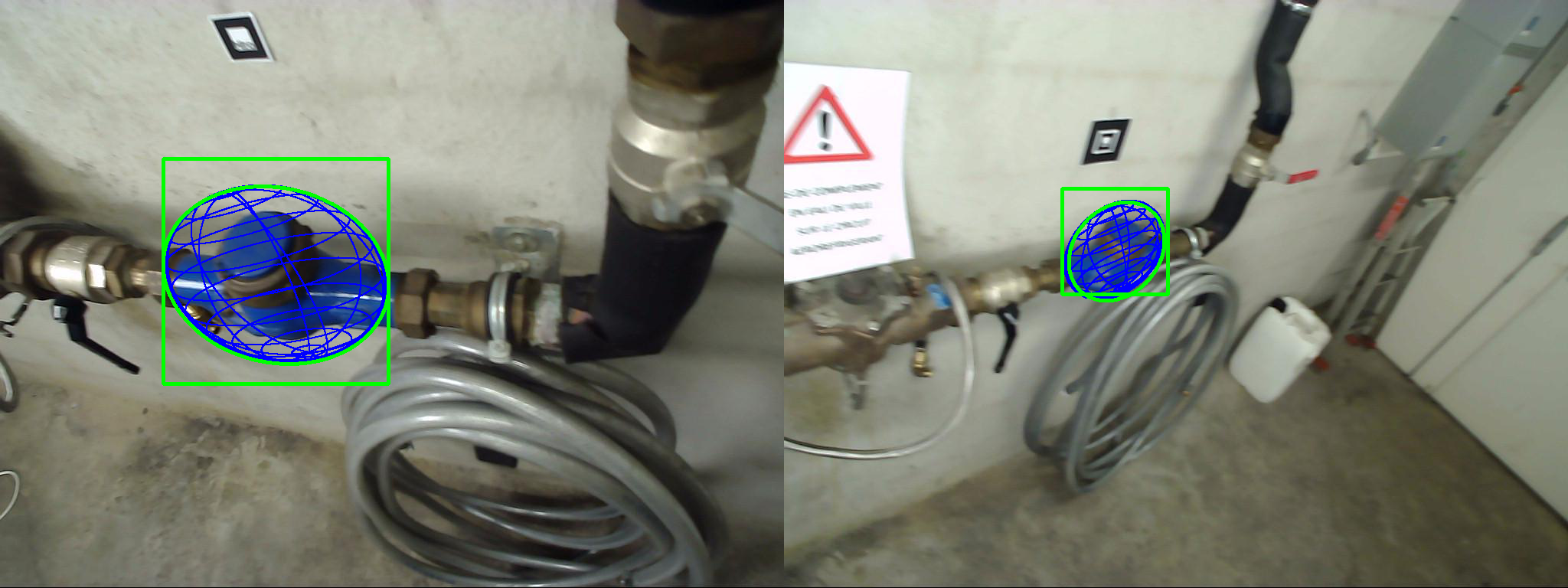}
    \caption{\textbf{Robustness to new viewpoints (position only):} Predicted ellipses \mzadd{(\textit{green})} and ground truth projections \mzadd{(\textit{blue})} of the ellipsoid in WatchPose (\textit{near} image on the left and \textit{far} image on the right). \mzadd{The green box shows the square detection box which contains the sub-image passed to the ellipse prediction network.}}
    \label{fig:WatchPose_results}
\end{figure}

\paragraph{Position estimation.}
We used the WatchPose dataset \cite{yang:hal-02735272}, which provides ten industrial scenes with images taken at different distances (\textit{near} at around 60cm and \textit{far} at around 1.4m). Unfortunately, only one object per scene can be used for localization, and thus, only the camera position was evaluated. We tested two scenarios: an easy one, where a subset of \textit{near} and \textit{far} images were used for training and testing, and a hard one, where training was done only on \textit{near} images and testing on \textit{far} images. Examples of predicted ellipses are available in Figure~\ref{fig:WatchPose_results}. The results in Table~\ref{tab:WatchPose_easy_and_hard} show the benefits offered by the the 3D-coherent ellipses, even in the \textit{far} case.

\begin{table}
  \scriptsize%
	\centering%
  \begin{tabu}
{|c|c|c|ccc|}
  \hline
      \multirow{2}{*}{Case} & \multirow{2}{*}{Method} & Median & \multicolumn{3}{c|}{Threshold} \\
      \cline{4-6}
      
       & & error (mm) & 5 cm & 10 cm & 15 cm \\
  \hline
\multirow{2}{*}{Easy} & Inscribed ell. & 84.11 & 27.87 & 57.38 & 77.05  \\
 & Ours & \textbf{26.63} & \textbf{77.27} & \textbf{97.72} & \textbf{100.0} \\

   \hline
\multirow{2}{*}{Hard} & Inscribed ell. & 120.31 & 14.61 & 38.20 & 66.29\\
& Ours & \textbf{54.12} & \textbf{40.74} & \textbf{81.48} & \textbf{88.88} \\

  \hline
  \end{tabu}%
  \caption{\textbf{Robustness to new viewpoints (position only):} Position errors obtained on WatchPose in the \textit{easy} and \textit{hard} cases (training and testing at mixed distances vs training only near the object and testing at larger distances).}
  \label{tab:WatchPose_easy_and_hard}
\end{table}

\begin{figure*}[ht]
\begin{subfigure}{.5\textwidth}
  \centering
  \includegraphics[width=\linewidth]{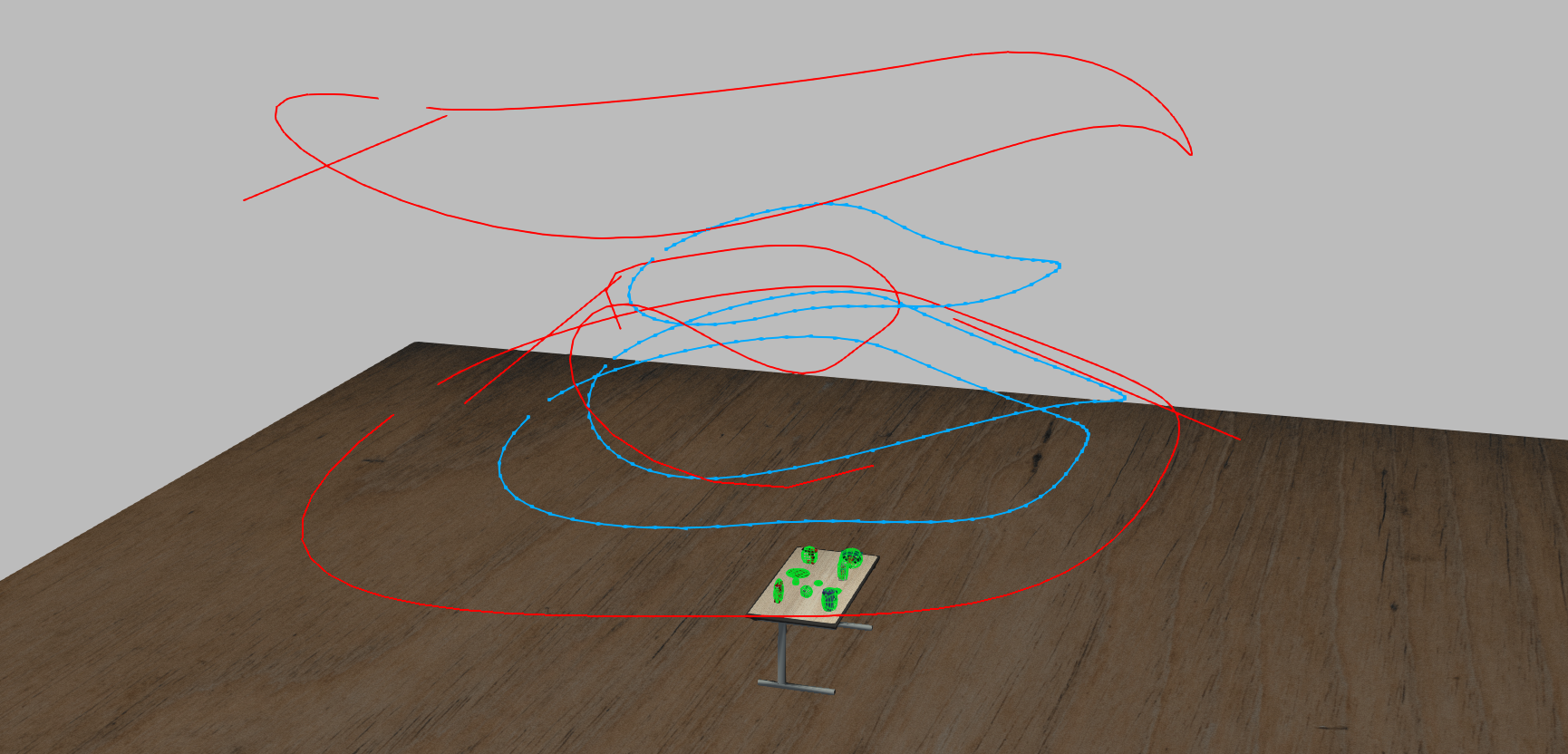}  
  \caption{Reconstructed scene model (the green ellipsoids) and trajectories used to generate the images (\textit{blue} for training and \textit{red} for testing).}
  \label{fig:Virtual_Scene_trajectories}
\end{subfigure}
\begin{subfigure}{.5\textwidth}
  \centering
  \includegraphics[width=\linewidth]{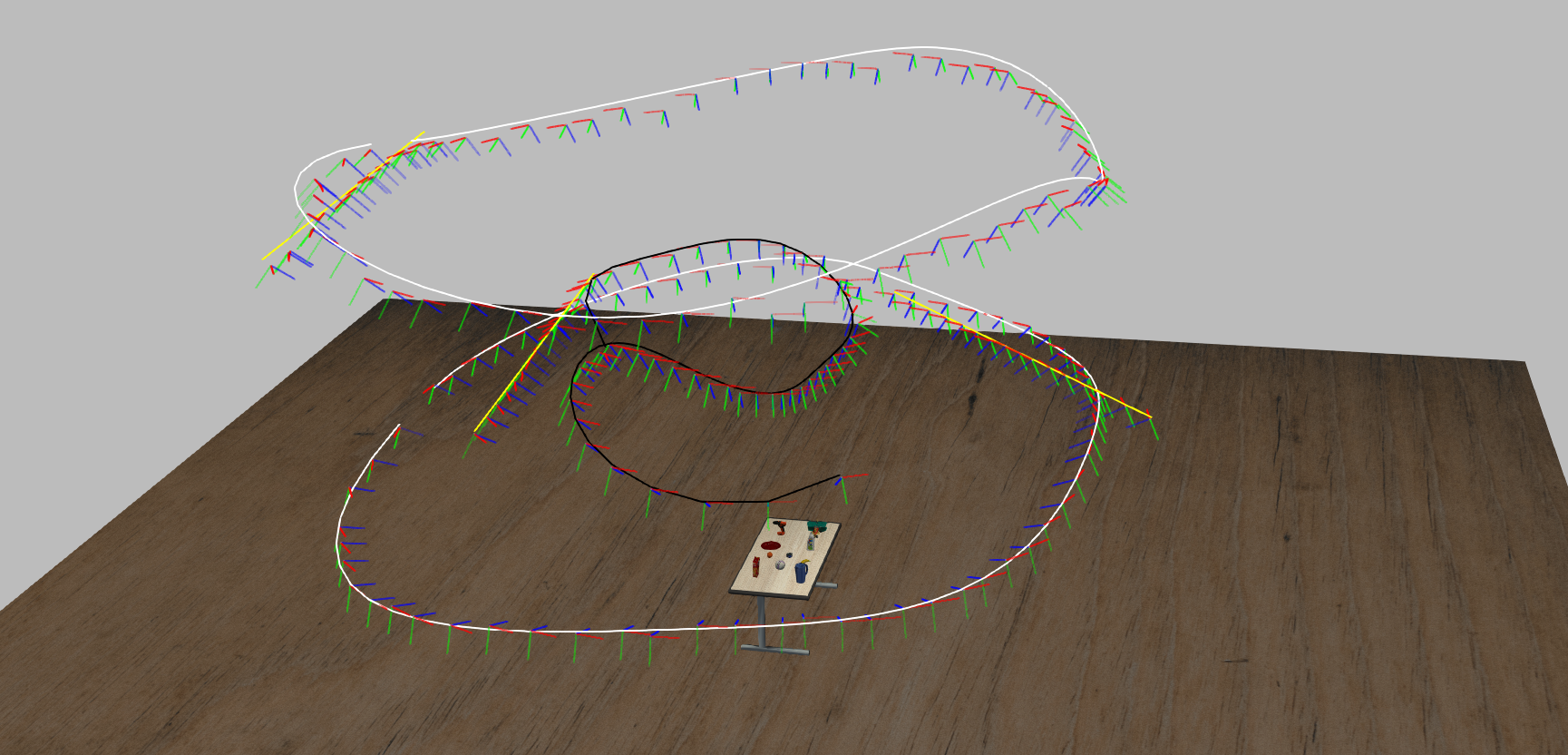}  
  \caption{Estimated camera poses obtained with our method. The ground truth camera trajectories are shown in black (test 1), yellow (test 2) and white (test 3).}
  \label{fig:Virtual_Scene_poses}
\end{subfigure}
\caption{\textbf{Robustness to new viewpoints (full pose):} Experiments on the virtual scene.}
\label{fig:Virtual_Scene}
\end{figure*}

\paragraph{Full pose estimation.}
In order to evaluate the robustness to new viewpoints of the full camera pose estimation, we created a synthetic dataset. This virtual scene is composed of ten objects taken from the YCB benchmark and rendered with Blender. This enabled us to completely control the camera viewpoints between training and testing. Figure~\ref{fig:Virtual_Scene} shows the scene with our reconstructed object models. The training images were generated from three camera trajectories taken at approximately \SI{4}{m} around the center of the scene, for a total of 192 images.
We then generated other camera trajectories for evaluation with more diverse viewpoints and more distant from the center of the scene (until \SI{8}{m} for the test case 3).

The obtained results are available in Table~\ref{tab:virtual_scene} and the estimated poses are visible in Figure~\ref{fig:Virtual_Scene_poses}. Our method achieves a very good accuracy on test cases 1 and 2, with a position error around \SI{5}{cm} and an orientation error around \SI{5}{\degree}. The errors are larger on test case 3, but stay acceptable given the large distance between the camera and the scene. Test cases 1 and 2 also show the benefits of using the predicted ellipses. In test case 3, the inscribed ellipses are slightly better. This can be explained by the fact that, at such distances, the objects appear very small in the image, which reduces the interest of predicting 3D-aware ellipses.

We also compared with the direct pose regression method PoseLSTM~\cite{DBLP:conf/iccv/WalchHLSHC17}. In contrast to our ellipse prediction module, this method has more difficulties to generalize to the new viewpoints of the test images. A noticeable difference which could explain the better generalization of our ellipse prediction module, is that, instead of using the whole image for prediction, it only takes as input a local patch around the objects. Indeed, this limits much more the change of appearance when the viewpoint changes.

 


\begin{table}[t]
  \scriptsize%
	\centering%
  \begin{tabu}{|c|c|ccc|}
  \hline
 
    \multicolumn{2}{|c|}{} & & & (Ours)\\
    \multicolumn{2}{|c|}{Method} & PoseLSTM & Inscribed & Predicted\\
    \multicolumn{2}{|c|}{} &   & ell. & ell.\\
   
  \hline
  \multirow{2}{*}{Test 1} & Pos. err. & 0.645 & 0.057 & \textbf{0.048} \\
                            & Rot error & 6.21 & 0.818 & \textbf{0.592} \\
  \hline
  \multirow{2}{*}{Test 2} & Pos. err. & 1.93 & 0.064&  \textbf{0.062} \\
                          & Rot error & 12.43 & 0.652 & \textbf{0.538}\\
  \hline
  \multirow{2}{*}{Test 3} & Pos. err. & 3.53 & \textbf{0.096} & 0.118 \\
                          & Rot error & 23.46 & \textbf{0.884} & 0.977\\
  \hline
  \end{tabu}%
  \caption{\textbf{Robustness to new viewpoints (full pose):} Mean position errors (in meters) and rotation errors (in degrees) on the three test cases of the scene.}
    \label{tab:virtual_scene}
\end{table}


\section{Analysis}

\subsection{Influence of the reconstructed ellipsoid}
\label{subseq:influence_of_the_reconstructed_ellipsoid}
\begin{figure}[H]
    \centering
    \includegraphics[width=0.9\linewidth]{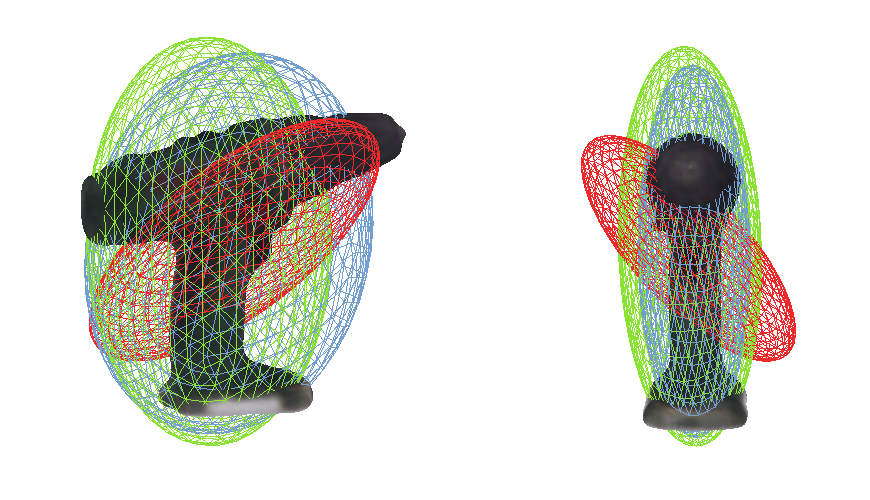}
    \caption{\textbf{Influence of the reconstructed ellipsoid:} The three different ellipsoids used in our experiment. Ellipsoid 1 from Table~\ref{tab:Linemod_different_ellipsoids} is in blue, ellipsoid 2 in green and ellipsoid 3 in red.}
    \label{fig:Linemod_different_ellipsoids}
\end{figure}

As our ellipsoidal models for objects depend on the ellipse observations used during the reconstruction, and, in particular, their viewpoints, it is important for our method to be able to learn the projection of more-or-less any ellipsoids, and not only the one which fits the best with the object. We verified it by repeating the experiment described in Section~\ref{subseq:Camera_position_estimation} on the \textit{driller} that we approximated with three different ellipsoids, shown in Figure~\ref{fig:Linemod_different_ellipsoids}. The results (in Table~\ref{tab:Linemod_different_ellipsoids}) are very similar for each model, which indicates that the choice of ellipsoid has no real influence.

\begin{figure}[H]
    \centering
    \includegraphics[width=\linewidth]{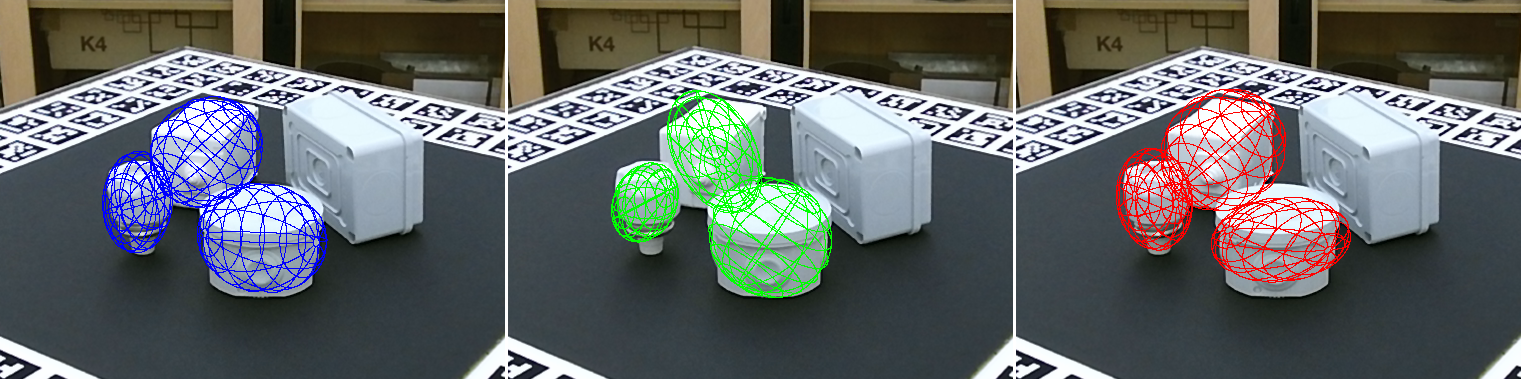}
    \caption{\mzadd{\textbf{Influence of the reconstructed ellipsoid:} Three different scene models. The blue ellipsoids were reconstructed from manual bounding box annotations and the green and red ones were manually deformed. Only three objects are used in the scene as it is sufficient to recover the full camera pose.}}
    \label{fig:TLESS_different_ellipsoids}
\end{figure}

\mzadd{We did a similar experiment on a scene with multiple objects, for which we are able to estimate the full 6D-pose of the camera. We reconstructed an ellipsoidal scene model from objects bounding box annotations in three images, and then, randomly deformed the ellipsoids into two other scene models (see Figure~\ref{fig:TLESS_different_ellipsoids}). We retrained the ellipse prediction network for each model. The results obtained are very similar in terms of pose accuracy for all the three scene models (see Table~\ref{tab:TLESS_different_ellipsoids}), which confirms again that our method does not strongly depend on the fitting accuracy between the ellipsoidal models and the real objects in 3D.}

\begin{table}[h]
  \scriptsize%
	\centering%
  \begin{tabu}{|@{\hspace{1mm}}c@{\hspace{1mm}}|@{\hspace{1mm}}c@{\hspace{1mm}}|@{\hspace{1mm}}c@{\hspace{1mm}}|@{\hspace{1mm}}c@{\hspace{1mm}}|}
  \hline
    
   Metric & Reprojection error & Position error & ADD\\
   Threshold & 5 pixels & 5 cm & 10\% of diam.\\
   \hline
Ellipsoid 1 & 96.31 & 95.85 & 85.71 \\
Ellipsoid 2 & 96.31 & 96.31 & 85.25 \\
Ellipsoid 3 & 98.62 & 95.85  & 84.79  \\
  \hline
  \end{tabu}%
  \caption{\textbf{Influence of the reconstructed ellipsoid:} Results obtained with different ellipsoidal models of the \textit{driller}.}  \label{tab:Linemod_different_ellipsoids}
\end{table}

\begin{table}[H]
    \centering
       \renewcommand*{\arraystretch}{1.2}
  \scriptsize%
	\centering%
  \begin{tabu}{|c|c|c|}
    \hline
         & Median position & Median orientation  \\
         & error (\SI{}{cm}) & error (\degree) \\
         \hline
        Ellipsoids 1 (blue) & 3.17 & 2.37 \\
        Ellipsoids 2 (green) & 3.10 & 2.44 \\
        Ellipsoids 3 (red) & 2.87 & 2.15 \\
        Ellipses inscribed & 4.81 & 3.58 \\
    \hline
    \end{tabu}
    \caption{\mzadd{\textbf{Influence of the reconstructed ellipsoid:} Median position and orientation errors obtained on TLESS (Fig.~\ref{fig:TLESS_different_ellipsoids}) for three different scene models. The three models provide similar results. The last line, where the pose is estimated using the inscribed ellipses, is given for comparison.}}
    \label{tab:TLESS_different_ellipsoids}
\end{table}

\subsection{Influence of the background on ellipse prediction}
\label{subsec:experiment_influence_background}

\begin{figure}[ht]
    \centering
    \includegraphics[width=\linewidth]{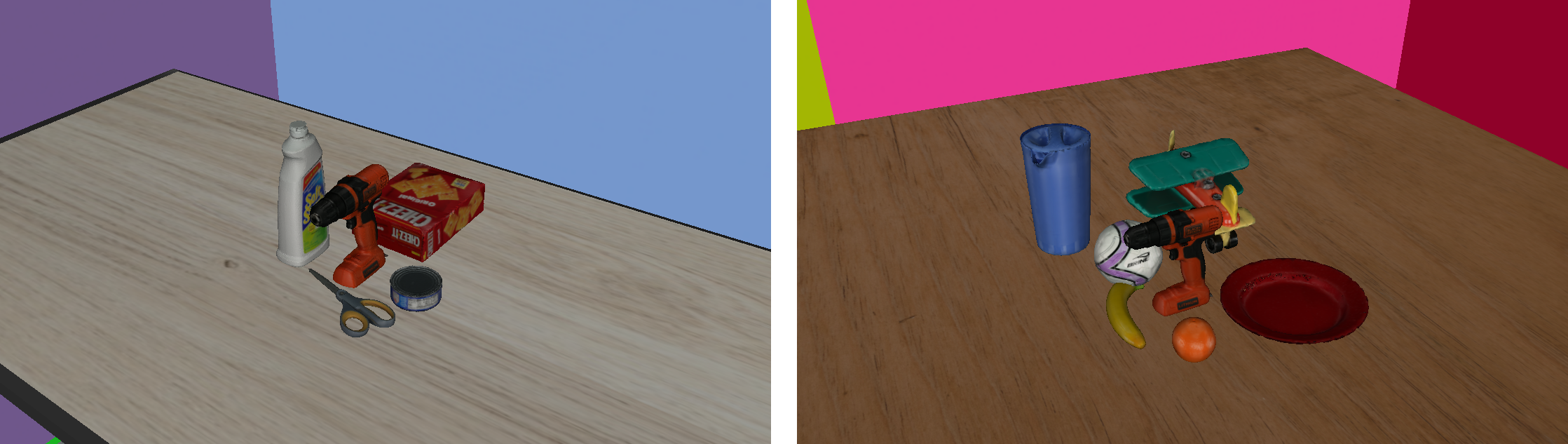}
    \caption{\textbf{Influence of the background:} The two synthetic scenes with the driller as central object in different environments, used to evaluate the influence of the background on the ellipse prediction.}
    \label{fig:Influence_of_background_scenes}
\end{figure}

\begin{figure}[ht]
\centering
\begin{subfigure}{.3\linewidth}
  \centering
  \includegraphics[width=\linewidth]{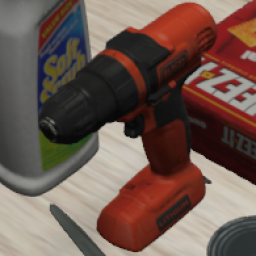}  
  \caption{No mask}
\end{subfigure}
\begin{subfigure}{.3\linewidth}
  \centering
  \includegraphics[width=\linewidth]{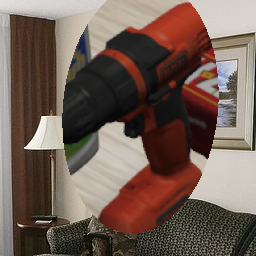}  
  \caption{Elliptic mask}
\end{subfigure}
\begin{subfigure}{.3\linewidth}
  \centering
  \includegraphics[width=\linewidth]{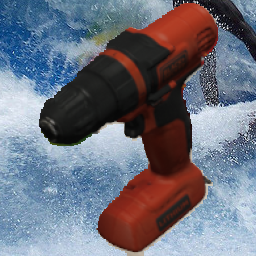}  
  \caption{GT mask}
\end{subfigure}
\caption{\textbf{Influence of the background:} The three masking strategies.}
\label{fig:Influence_of_background_masks}
\end{figure}

\begin{figure}[ht]
    \centering
    \includegraphics[width=\linewidth]{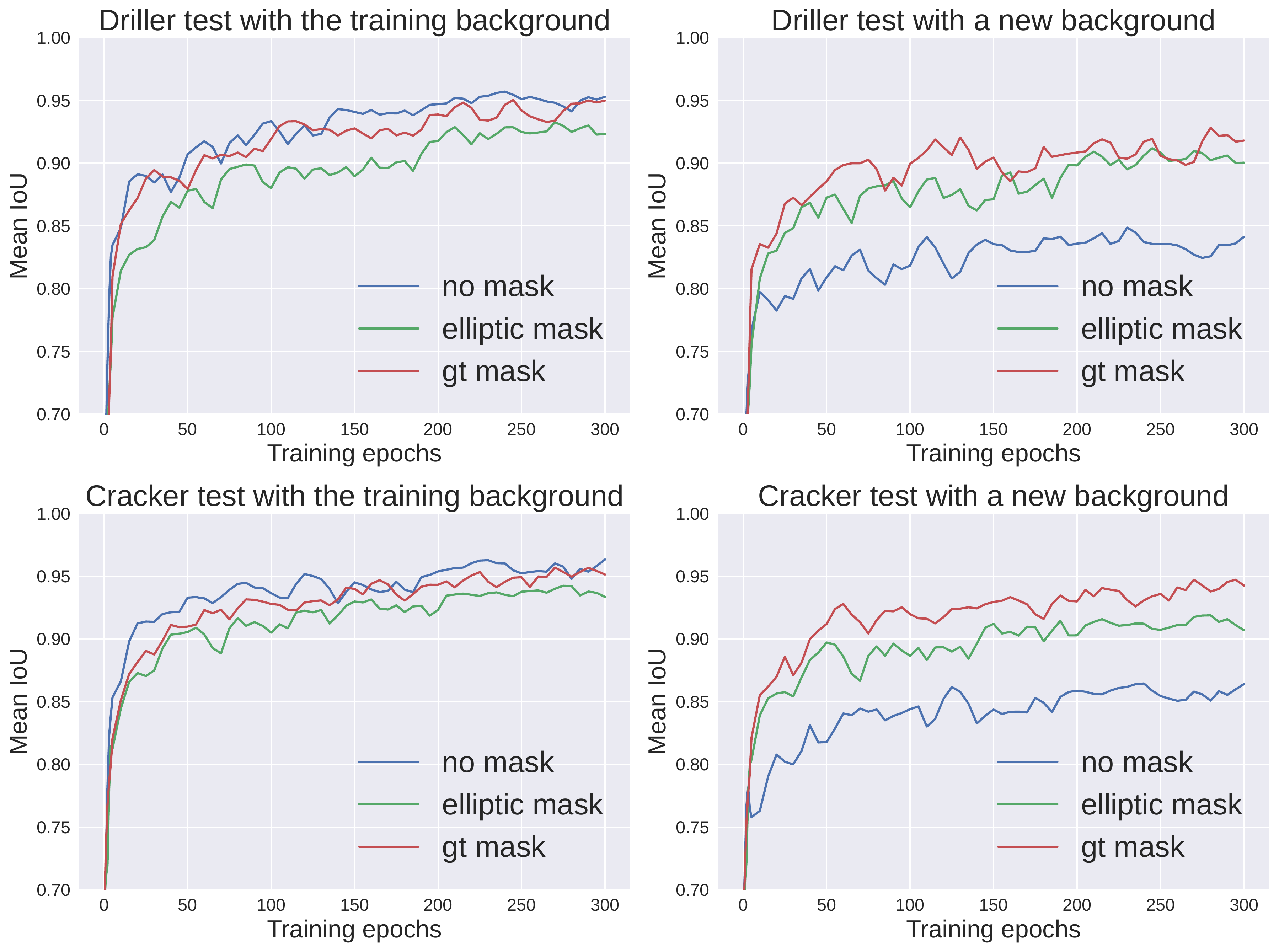}
    \caption{\textbf{Influence of the background:} Comparison of the three masking strategies in terms of mean IoUs obtained during 300 training epochs for the \textit{driller} and \textit{cracker} objects. The left column shows the results obtained on test images with the same background as in the training images. The right columns show the results obtained on test images with an unseen background.}
    \label{fig:Influence_of_background_results}
\end{figure}

As the ellipse prediction network takes a fixed-size crop image as input, a certain proportion of this image corresponds to background, depending on the object shape. In order to have a better understanding of how much this background interferes or contributes to the ellipse prediction, we created two synthetic scenes with the same central object but with different neighbouring objects and environments (see Figure~\ref{fig:Influence_of_background_scenes}). We then trained our ellipse prediction network on the first scene and tested it on both scenes. During training, we used three different masking strategies (Figure~\ref{fig:Influence_of_background_masks}):
\begin{enumerate}
    \item Our traditional method with a square crop image of the object.
    \item With an elliptic mask obtained by projecting the ellipsoidal model of the object and used to randomize the outside area.
    \item With a ground truth object mask used to randomize the background. 
\end{enumerate}
The randomized backgrounds were taken from COCO. Figure~\ref{fig:Influence_of_background_results} shows the evolution of the mean IoU obtained for the \textit{driller} and the \textit{cracker} on the two sets of test images at different times during the training. 

The results obtained on the test images with the same background as in training (left column of Figure~\ref{fig:Influence_of_background_results}) show relatively similar performances for all strategies. On both objects, the strategy without mask seems slightly better and  the elliptic mask slightly worse. This means that the network can benefit from the part of background visible in our crop images.

The second test, on the images of the same object in a different environment (right column), confirms this impression. This time, completely randomizing the background works the best whereas training with the original crop images gives the worst results. The method of elliptic masks is in-between and seems to still provide a good independence to the background with results only slightly lower than the ground truth masks. Nevertheless, the IoUs obtained in this second experiment stay relatively high (around 85\% without masks). This means that our network is still able to  predominantly use the object appearance in order to predict the ellipse. Of course, this will depend on the shape of the object and the proportion of visible background in the crop images. This link between the environment around the object and the predicted ellipse can thus be both beneficial or disadvantageous, depending on the usage. In our experiments, the scene remains static, which explains why we did not use any masking strategy. If a stronger independence to the background is required, using the elliptic mask strategy seems a good compromise as it does not require to know a precise 3D model of the object and takes advantage of the ellipsoid reconstruction step.

\subsection{Robustness to detection noise}
\begin{figure}
    \centering
    \includegraphics[width=\linewidth]{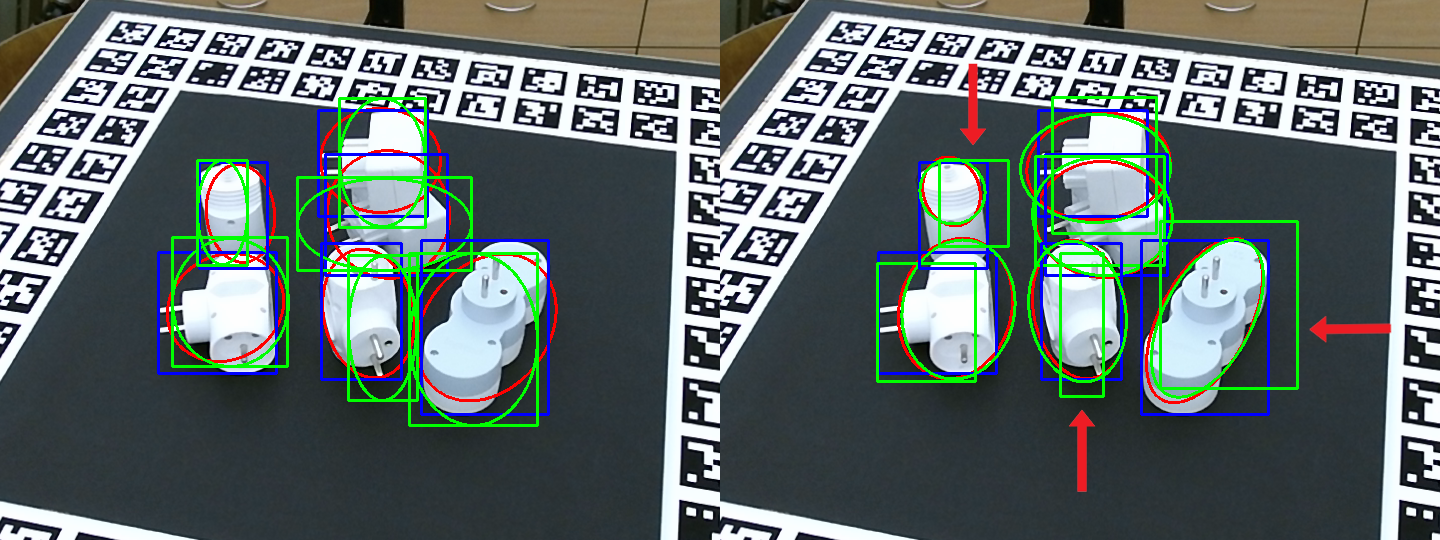}
    \caption{\textbf{Robustness to detection noise:} Inscribed ellipses (left) vs Predicted ellipses (right). Noisy BBs used for cropping and extracting the ellipses used for pose computation are in green. Ground truth projection of the ellipsoids are in red and ground truth objects BBs are in blue. Note that, despite noisy crops, the predicted ellipses fit much better to the ground truth projections.}
    \label{fig:tless_influence_of_noise_img}
\end{figure}

\begin{figure}
    \centering
    \includegraphics[width=\linewidth]{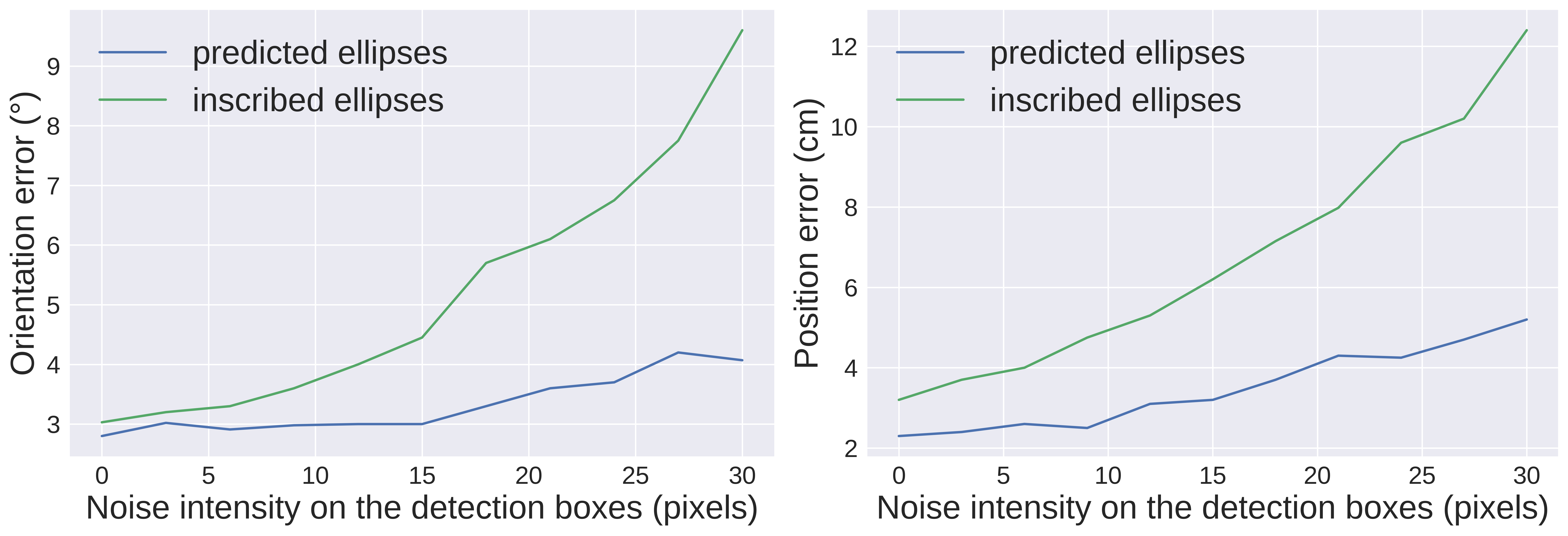}
    \caption{\mzadd{\textbf{Robustness to detection noise:} Influence of noisy BB detections on the estimated pose. Left: Orientation error (in degrees). Right: Position error (in cm). Both horizontal axes represent the half-range of the noisy shifts applied to the BBs.}}
    \label{fig:tless_influence_of_noise}
\end{figure}

An interesting point of our method is that the ellipse prediction module is quite robust to noisy detection boxes. We evaluated this ability on a scene with multiple objects taken from the T-LESS dataset~\cite{tless}, where we simulated noisy detections. We randomly shifted the corners defining each detection box.
Figures~\ref{fig:tless_influence_of_noise_img} and \ref{fig:tless_influence_of_noise} compare the influence of these noisy boxes on the inscribed ellipses and on the predicted ones.

On the one hand, it is easy to see that this spatial noise has a direct impact on the inscribed ellipses.
On the other hand, the predicted ellipses seem more robust. This is especially true for the objects marked with the arrows in Figure~\ref{fig:tless_influence_of_noise_img}. Even though the crop passed to the prediction network does not contain the whole object, the inferred ellipses are still correct. This robustness is mostly achieved thanks to our data augmentation which randomly shifts the image (equivalent to shifting the detection box before cropping).

\subsection{Comparison with the previous multi-bin loss}

We compared our new loss formulation with the previous multi-bin approach. We reused the 11 objects which were used in the experiment on 7-Scenes (Section~\ref{subseq:7-Scenes}). The results obtained on sequence 2 are summarized in Table~\ref{tab:new_loss_7-Scenes_objects}, in terms of Intersection-over-Union between the predicted ellipses and the ground truth ones. Our new loss outperforms the multi-bin loss on each object, which clearly shows the benefits offered by a more natural way of handling the angular discontinuity.

\begin{table}[t]
   \renewcommand*{\arraystretch}{1.2}
  \scriptsize%
	\centering%
  \begin{tabu}{|@{\hspace{2mm}}c@{\hspace{2mm}}|cc|cc|}
  \hline
  \multirow{3}{*}{Objects} & \multicolumn{2}{c|}{Multi-Bin loss} &  \multicolumn{2}{c|}{Sampling-based loss} \\
         \cline{2-5}
     & Mean  & \% & Mean & \% \\
     &  IoU &  IoU $>$ 0.8 & IoU & IoU $>$ 0.8 \\
    \hline
Tv (left)      &   0.912  &   1.0      &    \textbf{0.96}    &   1.0   \\
Tv (right)     &   0.897  &   1.0      &    \textbf{0.936}   &   1.0   \\
Xbox (left)    &   0.869  &   0.788    &    \textbf{0.943}   &   0.995 \\
Xbox (right)   &   0.854  &   0.8      &    \textbf{0.952}   &   0.997 \\
Chair (middle) &   0.906  &   0.948    &    \textbf{0.95}    &   1.0   \\
Chair (left)   &   0.888  &   0.802    &    \textbf{0.902}   &   0.831 \\
Chair (right)  &   0.847  &   0.7      &    \textbf{0.873}   &   0.837 \\
Chess clock    &   0.908  &   0.952    &    \textbf{0.936}   &   0.996 \\
Video Games    &   0.946  &   1.0      &    \textbf{0.945}   &   1.0   \\
Interrupter    &   0.918  &   0.997    &    \textbf{0.935}   &   1.0   \\
Gamepad        &   0.93   &   0.96     &    \textbf{0.941}   &   1.0   \\
  \hline
  \end{tabu}
  \caption{\textbf{Comparison with the previous multi-bin loss:} Mean IoU scores of the predicted ellipses obtained with the multi-bin loss and with our new loss based on implicit function sampling. The objects are those used in our experiment on 7-Scenes (Section~\ref{subseq:7-Scenes}).}
    \label{tab:new_loss_7-Scenes_objects}
\end{table}

\subsection{\mzadd{Comparison of different embedding functions}}
\label{subsec:analysis_embedding_functions}

\mzadd{We analyze here the performance of different embedding functions used in the loss of our ellipse prediction network.
More precisely, we only changed the form of the central matrix in Equation~\ref{eq:our_implicit_function}, which is responsible for integrating the ellipse axes. The results in Table~\ref{tab:analysis_embedding_functions} were obtained on two objects: the synthetic driller, already used in the experiment described in subsection~\ref{subsec:experiment_influence_background}, and the real driller from LINEMOD.}

\begin{figure}[ht]
    \centering
    \includegraphics[width=\linewidth]{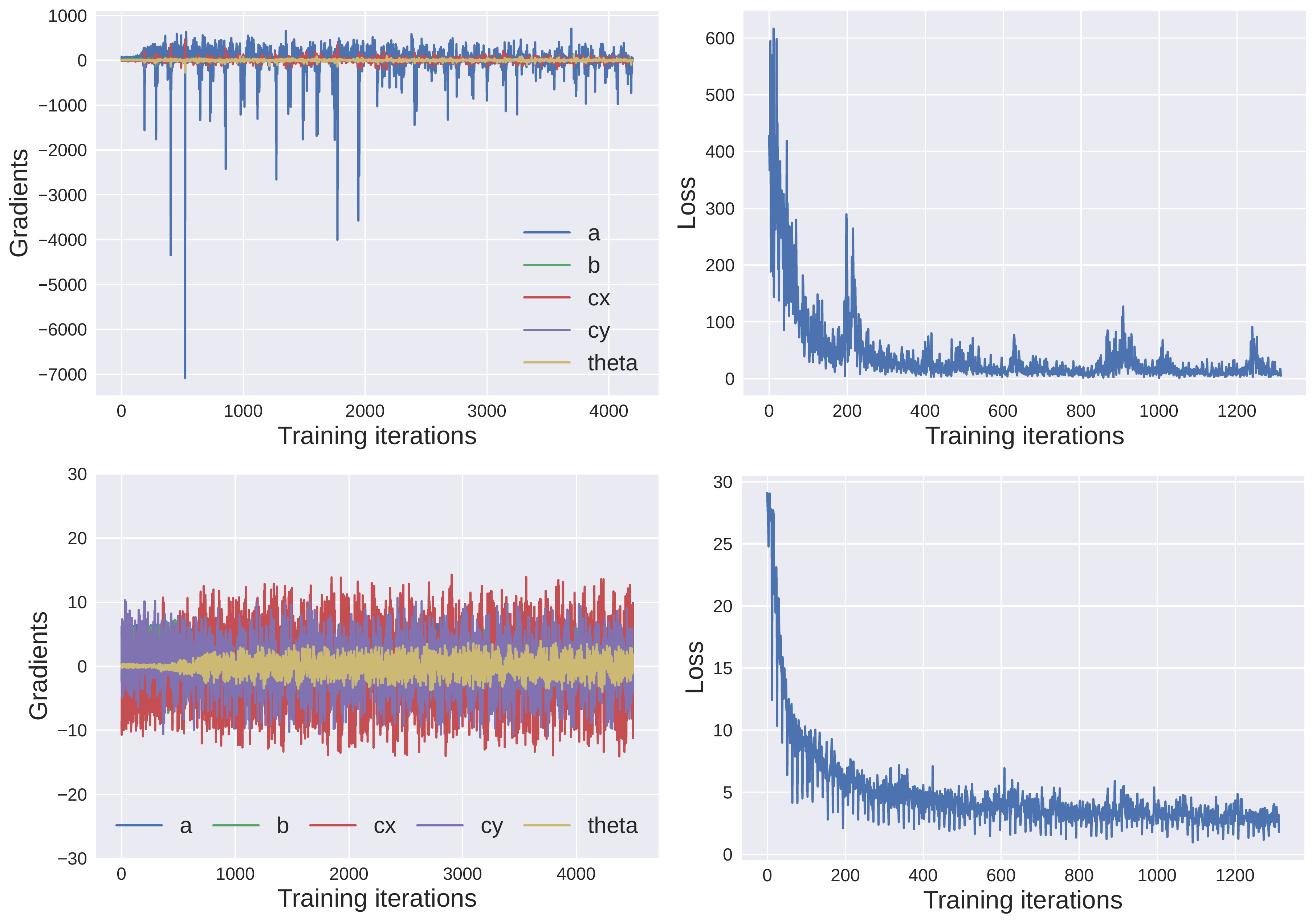}
    \caption{Evolution of the gradient wrt. the ellipse parameters and the loss during training. \textit{Top row}: The quadratic equation of an ellipse is directly used as intermediate function representation. \textit{Bottom row}: Our proposed function~(\ref{eq:our_implicit_function}) is used. Note the difference in scale between the two rows.}
    \label{fig:New_loss_inverse_analysis}
\end{figure}

\mzadd{They show that the simplified version, with only $[\alpha, \beta]$ on the diagonal, provides the best results. The lower performances of the other expressions are probably caused by numerical instability encountered during training and which can be observed in Figure~\ref{fig:New_loss_inverse_analysis}. In particular, the huge gradient values can be explained by the form of the derivative of the expressions on the diagonal of the central matrix, $[\frac{-2}{\alpha^3}, \frac{-2}{\beta^3}]$ which can become huge when $\alpha$ and $\beta$ are small. In our case, $\alpha$ and $\beta$ are normalized between 0 and 1.
}

\begin{table}[]
    \centering
       \renewcommand*{\arraystretch}{1.2}
  \scriptsize%
	\centering%
  \begin{tabu}{|c|c|c|c|c|}
    \hline
        central & \multicolumn{2}{c|}{Synthetic driller} & \multicolumn{2}{c|}{LINEMOD driller} \\
        \cline{2-5}
        
        matrix & Mean & \% IoU & Mean & \% IoU\\
        form & IoU & $> 0.8$ & IoU & $> 0.8$\\
        \hline
        $\left[\begin{array}{cc}
            \frac{1}{\alpha^2} & 0 \\
            0 & \frac{1}{\beta^2} 
        \end{array}\right]$
        & 0.873 & 0.86 & 0.866 & 0.90\\

        $\left[\begin{array}{cc}
            \frac{1}{\alpha} & 0 \\
            0 & \frac{1}{\beta} 
        \end{array}\right]$
        & 0.910 & 0.97 & 0.903 & 0.98\\
            
        $\left[\begin{array}{cc}
            \alpha^2 & 0 \\
            0 & \beta^2 
        \end{array}\right]$
        & 0.919 & \textbf{0.98} & 0.903 & 0.97\\
        
        $\left[\begin{array}{cc}
            \alpha & 0 \\
            0 & \beta
        \end{array}\right]$
        & \textbf{0.921} &\textbf{ 0.98} & \textbf{0.930} & \textbf{0.99}\\
    \hline
    \end{tabu}
    \caption{\mzadd{Mean IoU and percentage of predicted ellipses with an IoU greater than 0.8 with the ground truth ellipse for different embedding functions. The evaluation was performed for two objects: the \textit{orange synthetic driller} (Fig.~\ref{fig:TLESS_different_ellipsoids}), and the \textit{LINEMOD driller} (Fig.~\ref{fig:Linemod_predictions_for_some_objects}).}}
    \label{tab:analysis_embedding_functions}
\end{table}

\section{\mzadd{Discussion: class-level vs. instance-level}}

\mzadd{
In our method, ellipse prediction is done at instance-level while object detection is performed at class-level. One might ask if a single multi-class ellipse prediction network is conceivable or whether an end-to-end training for both object detection and ellipse predictions is possible. Actually, in our case, two instances of the same object do not necessarily share the same ellipsoidal representation, as these are simply obtained from multi-view reconstruction. That is why predicting an ellipse coherently with the ellipsoidal 3D model of a specific object requires instance-level awareness, whereas typical object detection networks work at class-level.
Training a network to differentiate two instances of the same object is challenging and likely to lead to erroneous predictions when neither the visual aspect of the objects nor the background provide enough information. Wrong object instances associations would directly degrade the estimated camera pose.
Instead, we detect objects at class-level, predict multiple ellipses hypotheses and disambiguate them in the RANSAC loop, which is more likely to discard wrong correspondences.}\mzadd{Also, this provides more flexibility for the detection part. In fact, any existing pre-trained network for object detection can be used (YOLO~\cite{RedmonF17}, Faster R-CNN~\cite{FasterRCNN}, DETR~\cite{DETR}, ...).
}



\section{Conclusion}

In this paper, we proposed a method for object-based camera pose estimation which does not require an accurate model of the scene. Its main component is a 3D-aware ellipse prediction network with an improved loss. By learning from different viewpoints, the network is able to map the object appearance to ellipse parameters which are coherent with the projection of the object ellipsoidal abstraction, and thus, improves the estimated camera pose. Three key aspects of the method are its good invariance to the chosen ellipsoidal models, its robustness to variance in the box detection boundaries and its minimal amount of manual annotations required, making the method of large practical interest.
\mzadd{While, the proposed method already provides poses with a good accuracy, adding a camera pose refinement step is an interesting direction to explore in future works. In particular, the question of establishing a cost between two ellipses (detection vs. reprojection) arises.}

\newcolumntype{C}{>{\centering\arraybackslash}m{.3\textwidth}}
\begin{table*}\sffamily
\begin{tabular}{@{\hspace{0mm}}c@{\hspace{0mm}}c@{\hspace{1mm}}c@{\hspace{1mm}}c@{\hspace{1mm}}}

\toprule
 & All frames & At least 2 detections & At least 3 detections \\ 
\midrule
Nb. frames & 1000 & 960 & 828 \\

\multirow{2}{*}{Seq 2} &
\includegraphics[width=.3\textwidth]{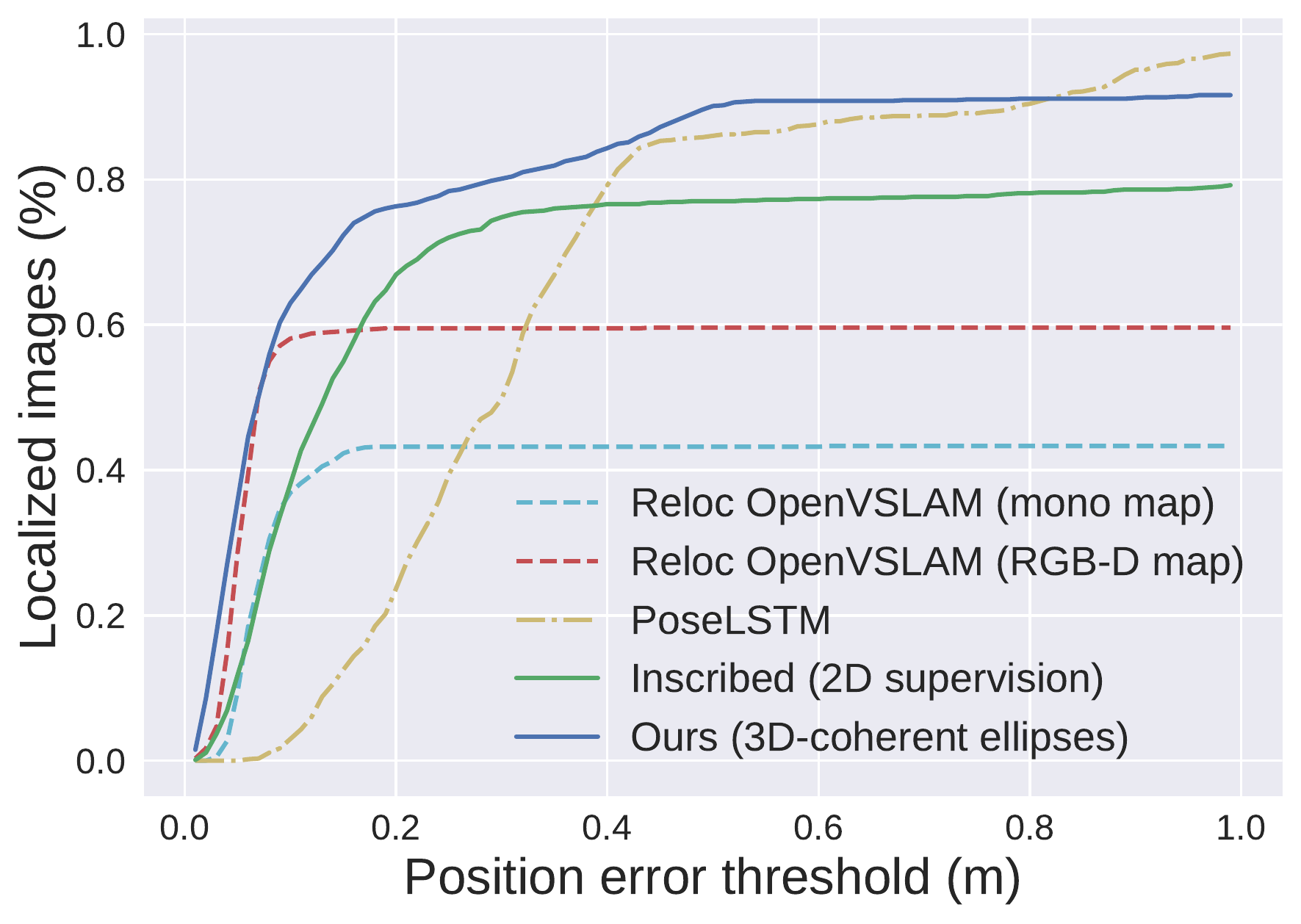}  &
\includegraphics[width=.3\textwidth]{position_acc_seq_seq_02_at_least_2-eps-converted-to.pdf}  &
\includegraphics[width=.3\textwidth]{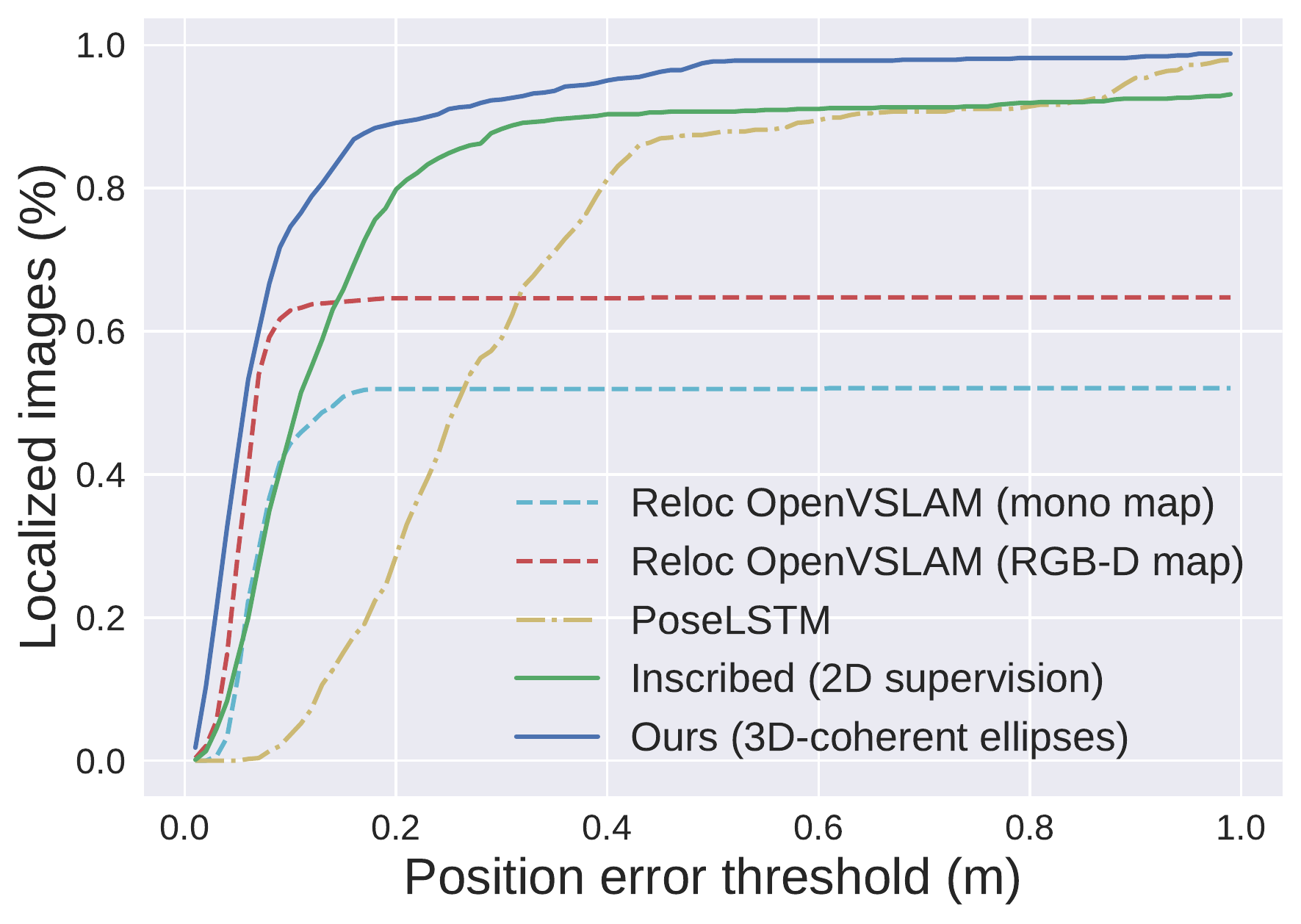}  
\\ 
&
\includegraphics[width=.3\textwidth]{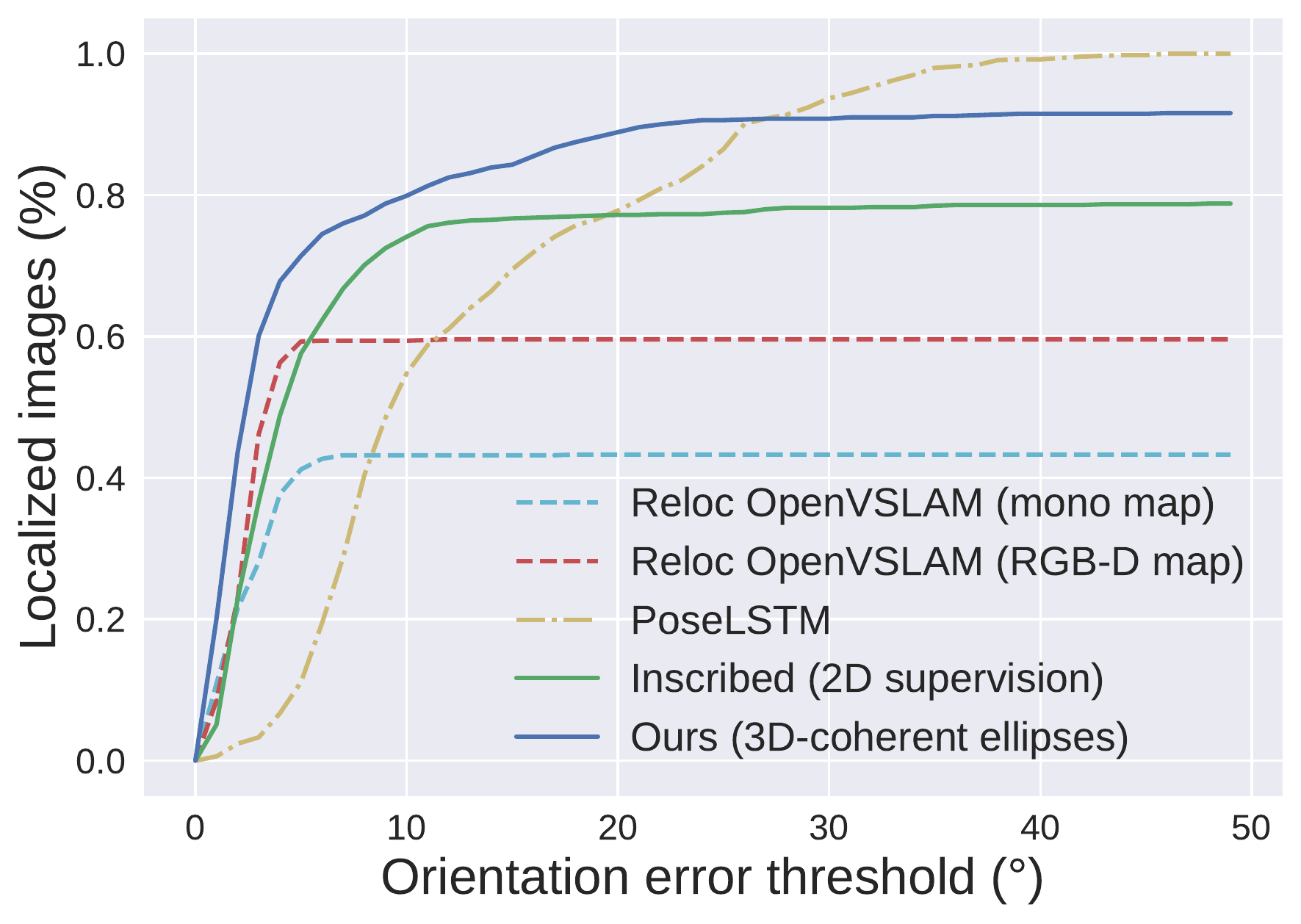}  &
\includegraphics[width=.3\textwidth]{orientation_acc_seq_seq_02_at_least_2-eps-converted-to.pdf}  &
\includegraphics[width=.3\textwidth]{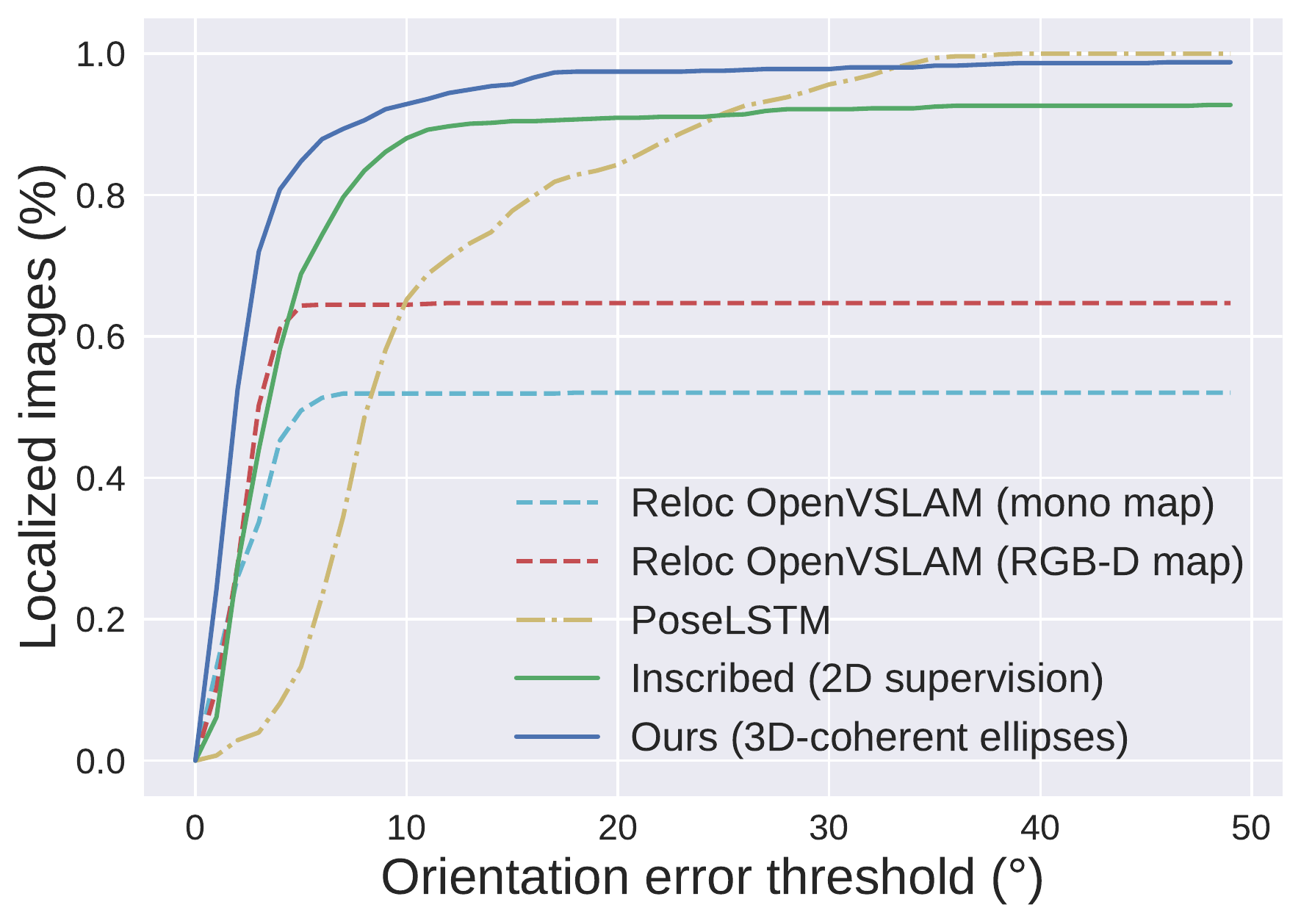}
\\
\midrule

Nb. frames & 1000 & 965 & 867 \\
\multirow{2}{*}{Seq 3} &
\includegraphics[width=.3\textwidth]{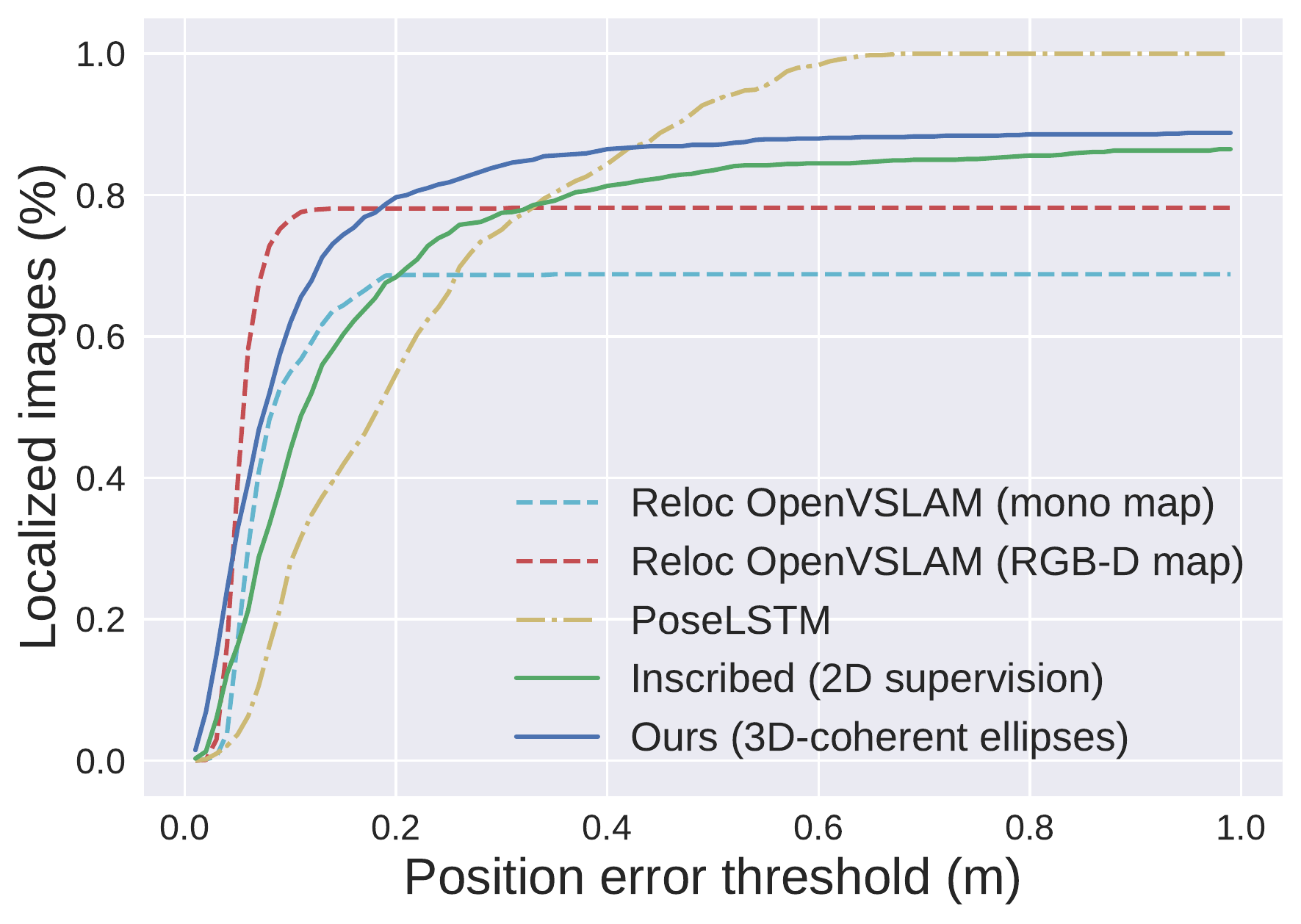}  &
\includegraphics[width=.3\textwidth]{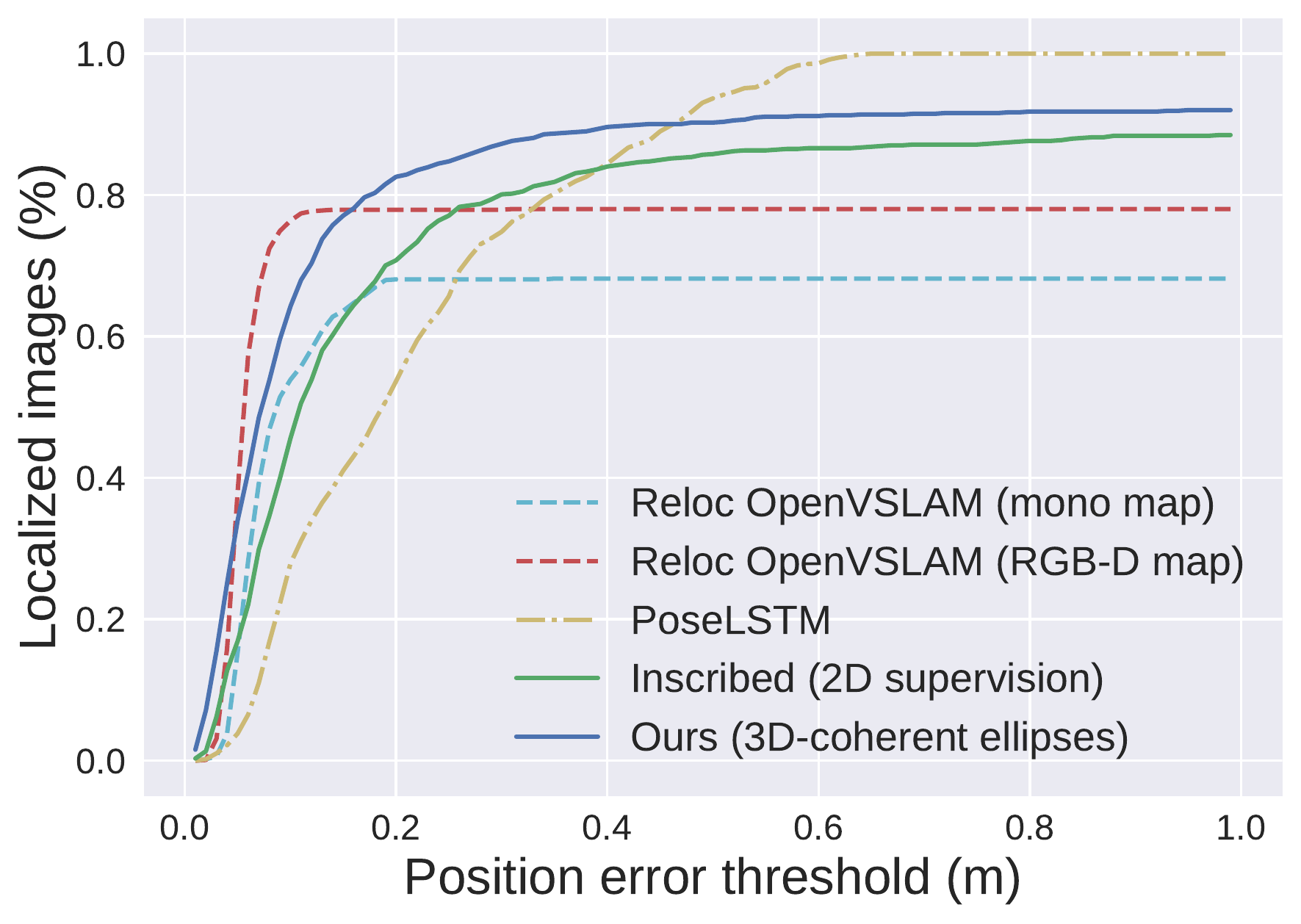}  &
\includegraphics[width=.3\textwidth]{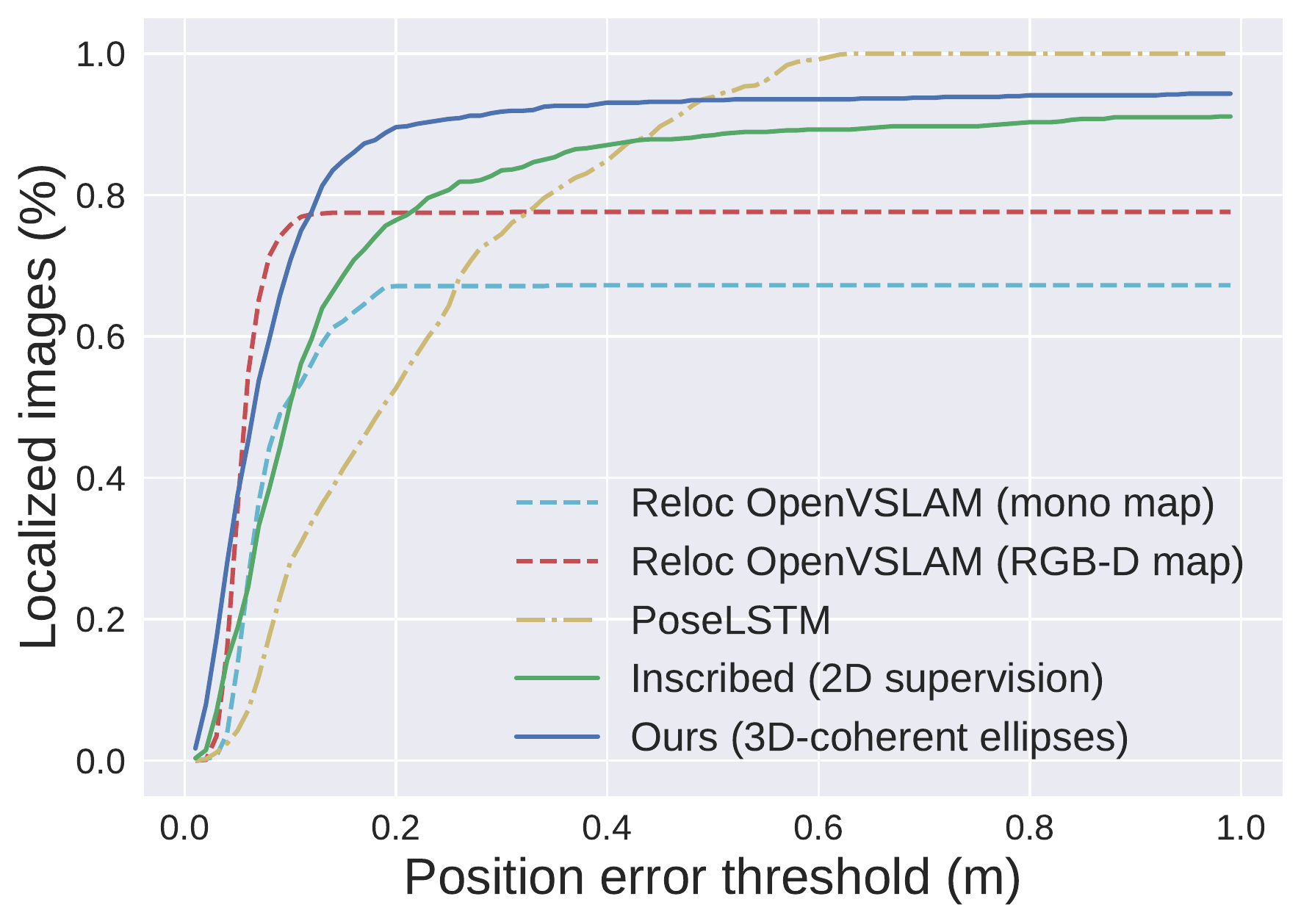}  
\\ 
&
\includegraphics[width=.3\textwidth]{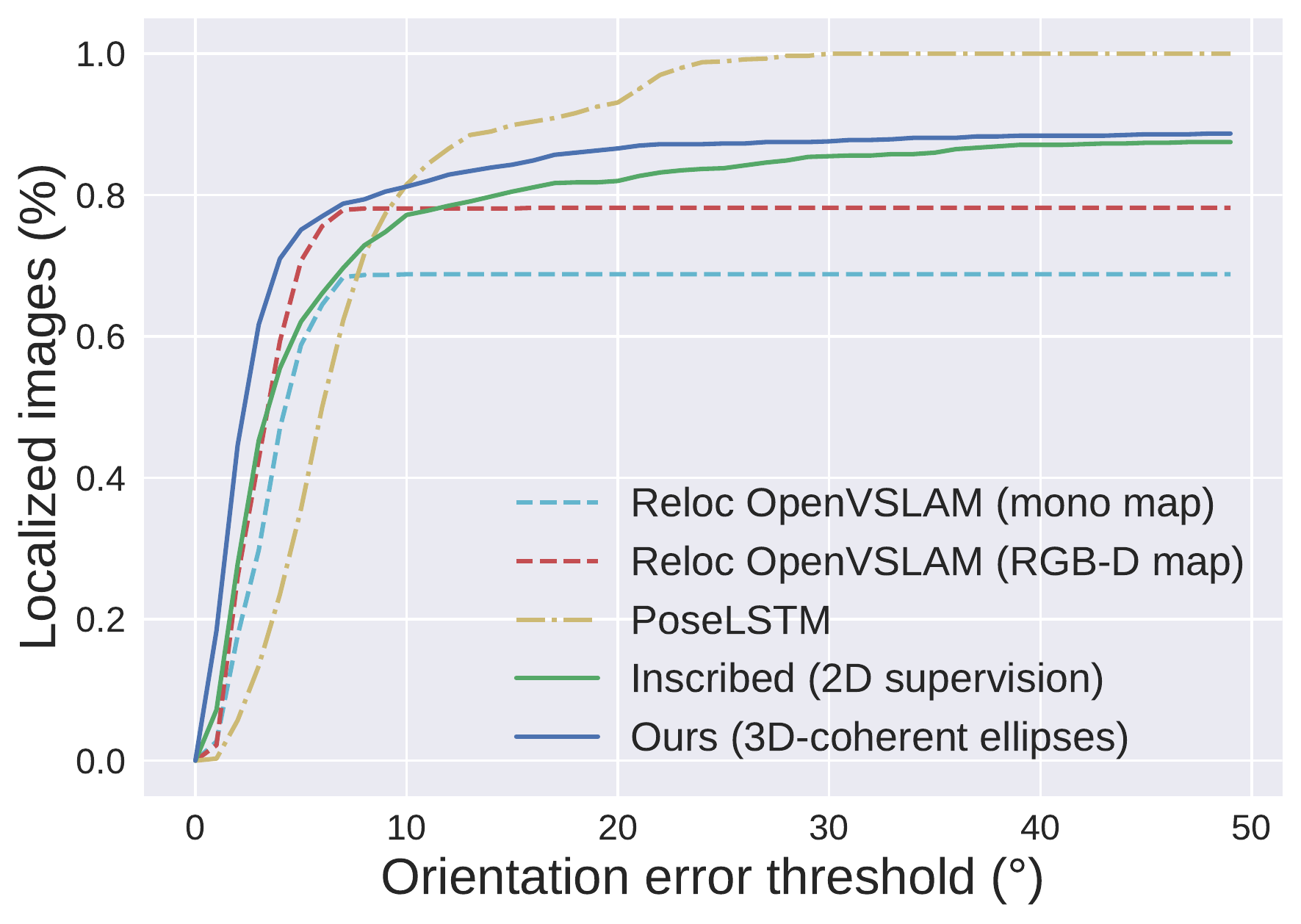}  &
\includegraphics[width=.3\textwidth]{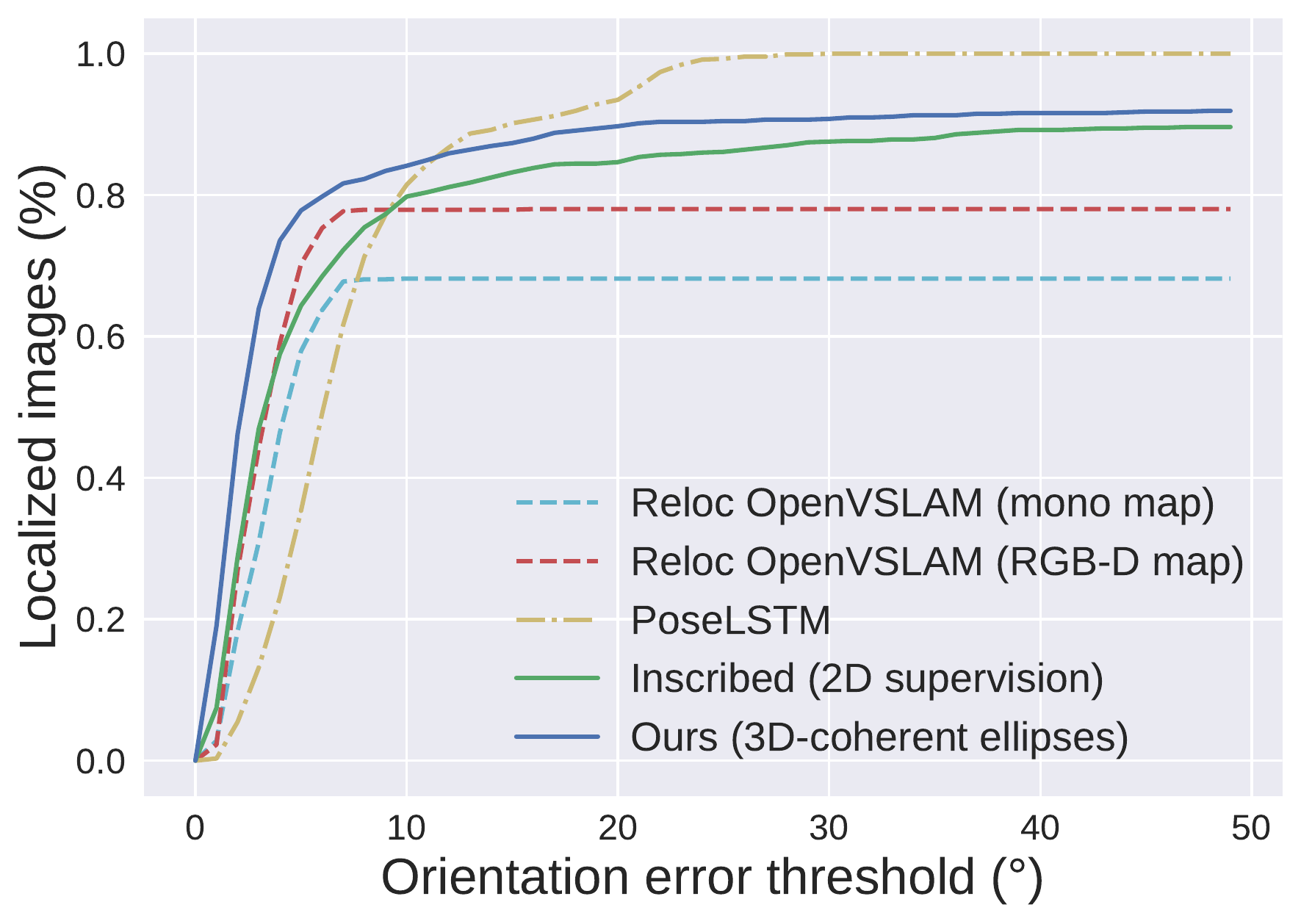}  &
\includegraphics[width=.3\textwidth]{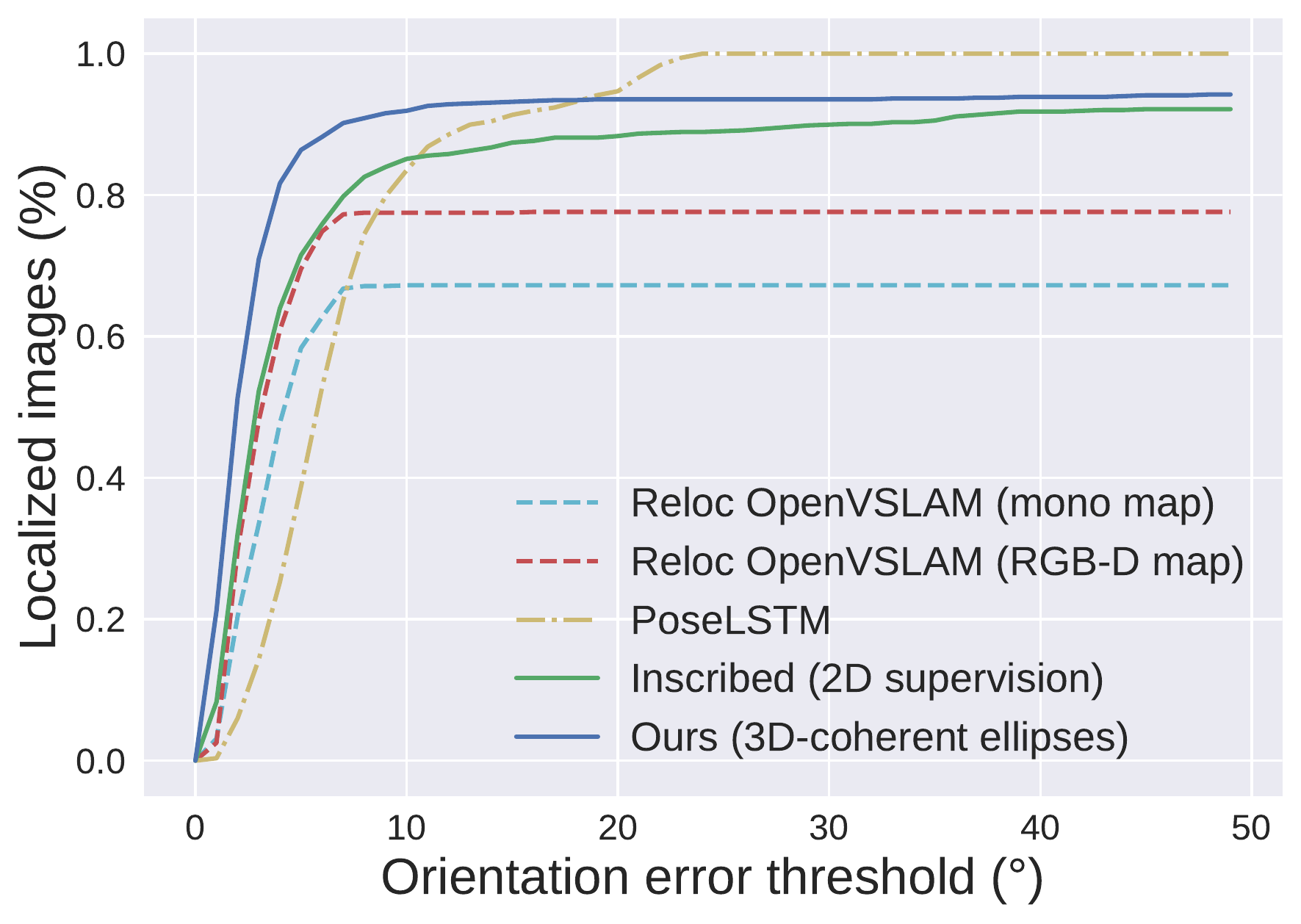}  
\\ \midrule

Nb. frames & 1000 & 973 & 930 \\
\multirow{2}{*}{Seq 5} &
\includegraphics[width=.3\textwidth]{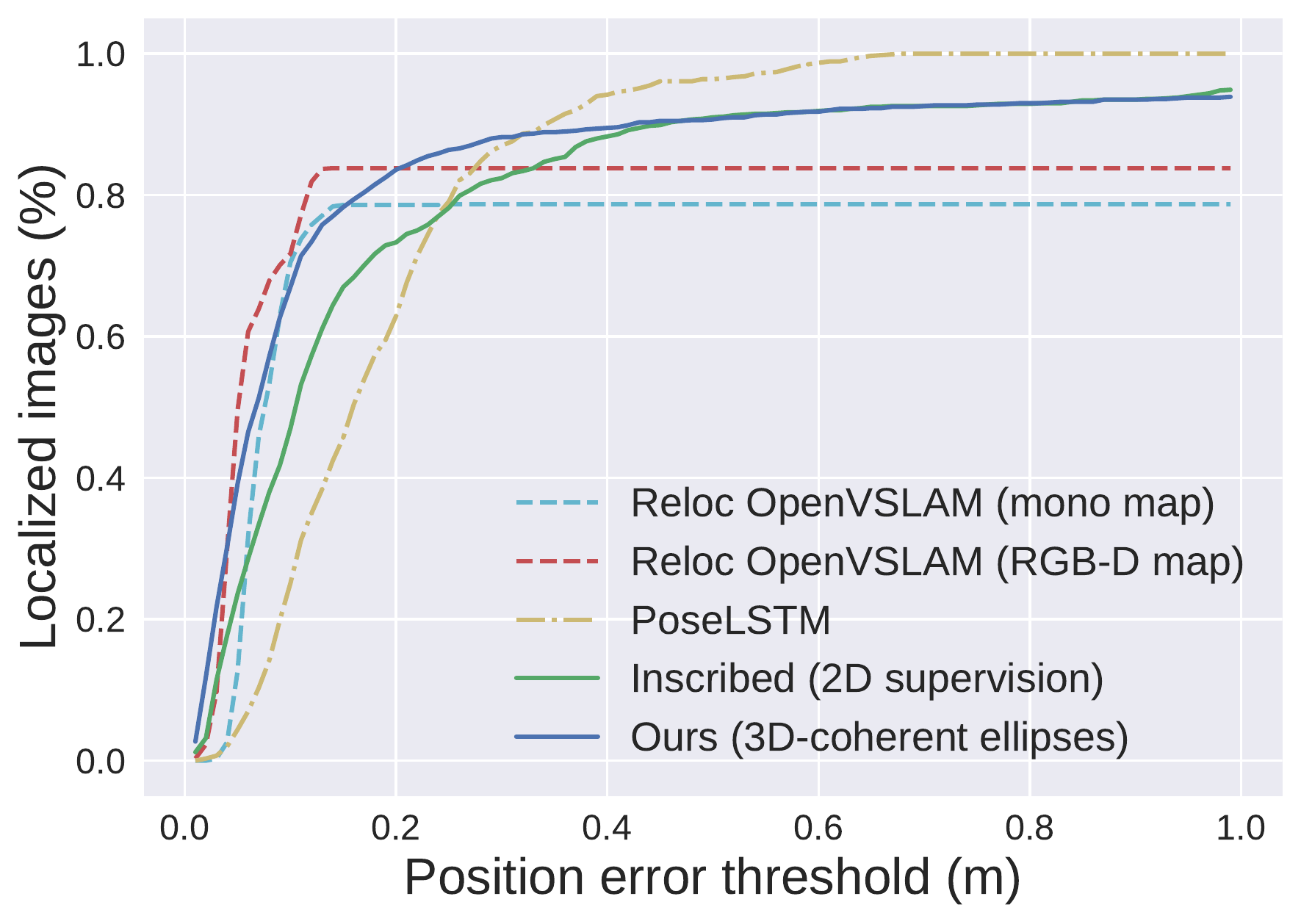}  &
\includegraphics[width=.3\textwidth]{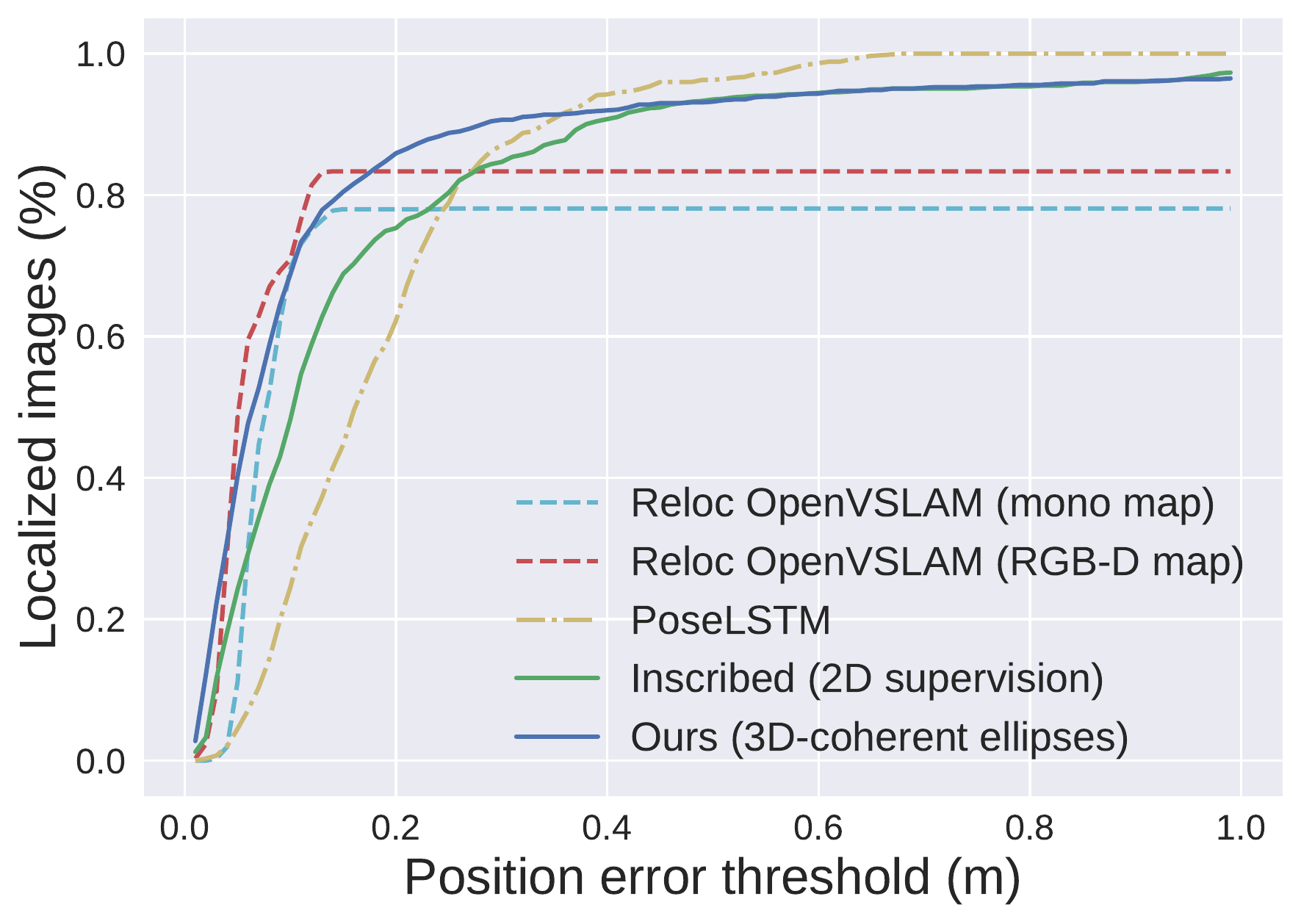}  &
\includegraphics[width=.3\textwidth]{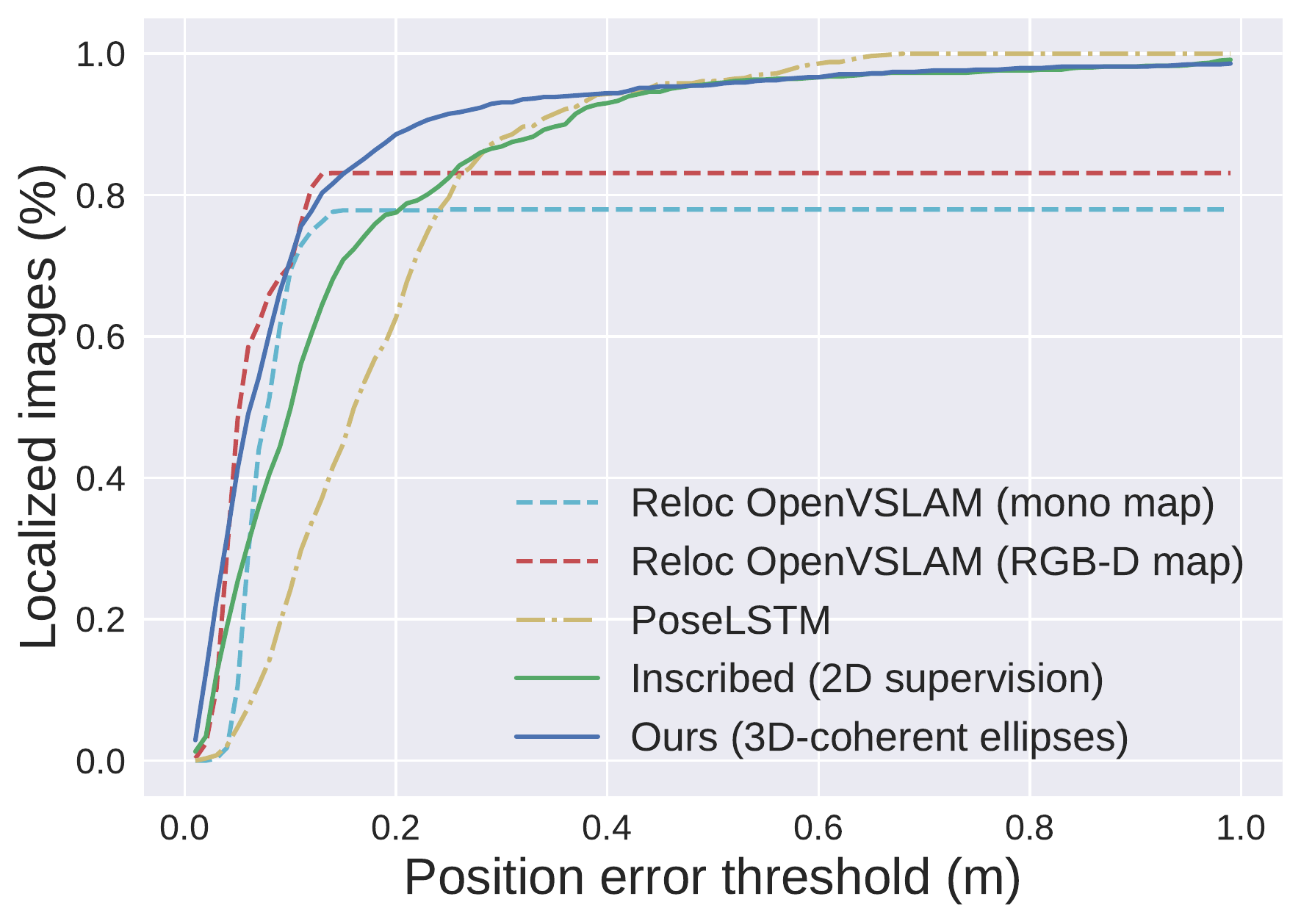}  
\\ 
&
\includegraphics[width=.3\textwidth]{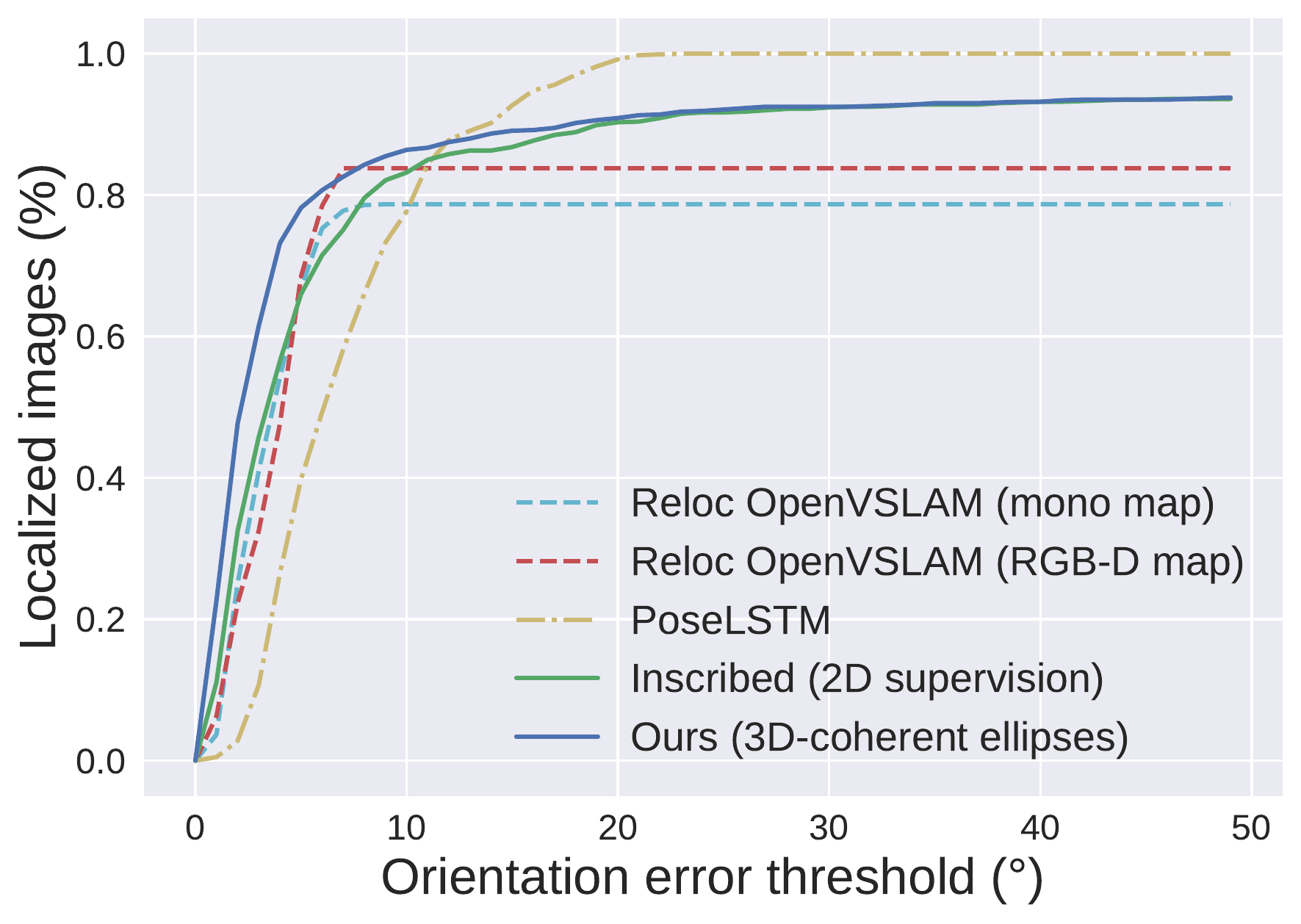}  &
\includegraphics[width=.3\textwidth]{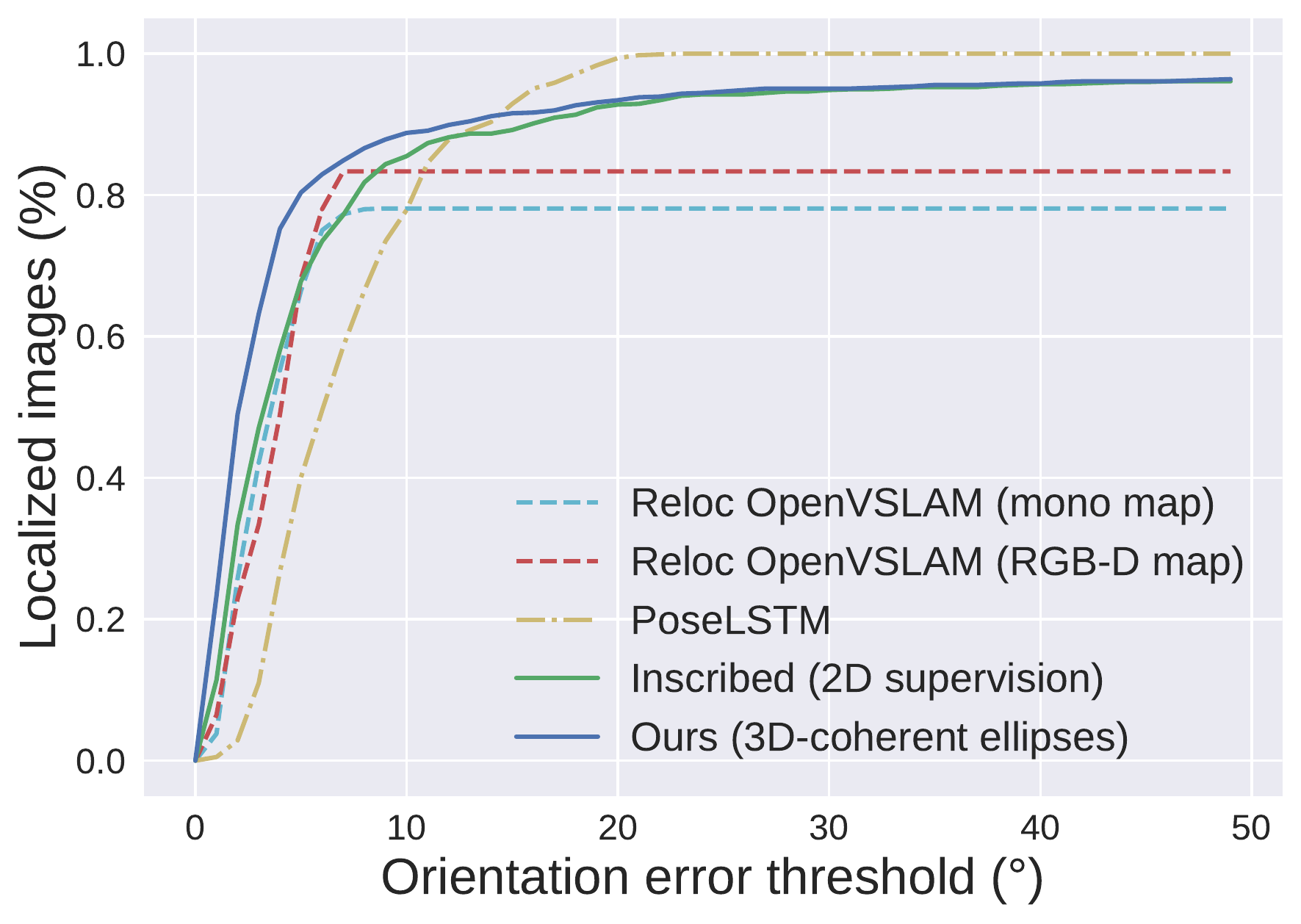}  &
\includegraphics[width=.3\textwidth]{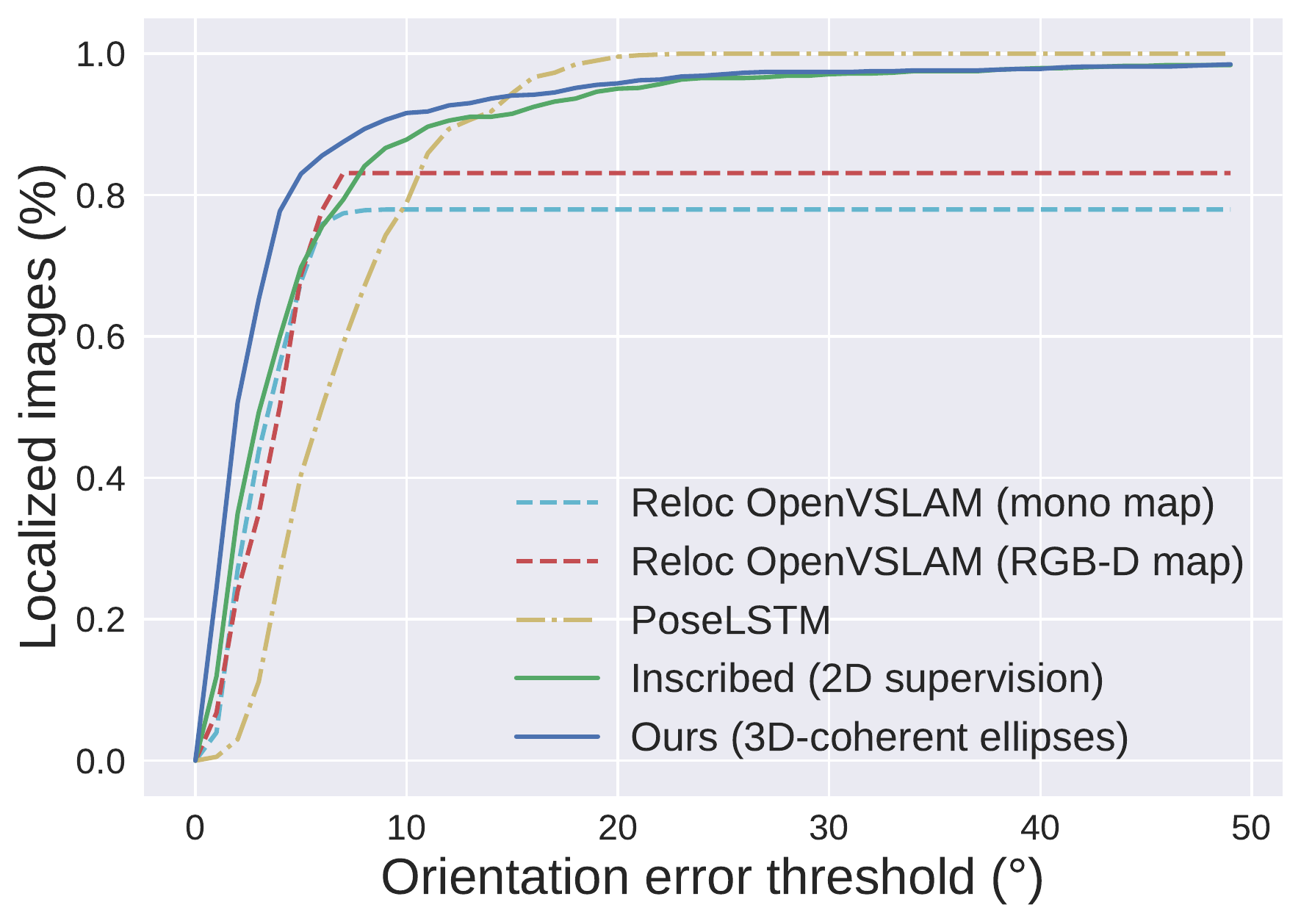}  
\\ \bottomrule

\end{tabular}
\caption{\textbf{Full camera pose estimation:} Proportion of correctly localized frames wrt. an error threshold on the \textit{Chess} scene. The columns represent different subsets of frames, according to the number of detected objects.}
\label{tab:7-Scenes_accuracies_big_table}
\end{table*}


%
%

\bibliographystyle{spmpsci}      
\bibliography{main}   


\end{document}